\documentclass[twocolumn, 10pt]{IEEEtran}
\usepackage{graphicx}
\usepackage{subcaption}
\usepackage{svg}
\usepackage{amssymb}
\usepackage{amsmath}
\usepackage{mathtools}
\usepackage{dsfont}
\usepackage{cite}
\usepackage{stfloats}
\usepackage{psfrag}
\usepackage[mathscr]{euscript}
\usepackage{acronym}  % make an acronym
\usepackage{booktabs}
\usepackage{bbm}
\usepackage{algorithmic}
\usepackage{algorithm}

\usepackage{float}
\usepackage{balance}

\acrodef{CCDF}{complementary cumulative distribution function}
\acrodef{CF}{characteristic function}
\acrodef{PPP}{Poisson point processe}
\acrodef{RV}{random variable}
\acrodef{i.i.d.}{independent and identically distributed}
\acrodef{PDF}{probability distribution function}
\acrodef{CDF}{cumulative distribution function}
\acrodef{ch.f.}{characteristic function}
\acrodef{AWGN}{additive white Gaussian noise}
\acrodef{SNR}{signal-to-noise ratio}
\acrodef{LRT}{likelihood ratio test}
\acrodef{DRT}{distance ratio test}
\acrodef{GLRT}{generalized likelihood ratio test}
\acrodef{CRLB}{Cram\'{e}r-Rao lower bound}
\acrodef{CRB}{Cram\'{e}r-Rao bound}
\acrodef{ZZLB}{Ziv-Zakai lower bound}
\acrodef{ZZB}{Ziv-Zakai bound}
\acrodef{LOS}{line-of-sight}
\acrodef{ToF}{time-of-flight}
\acrodef{NLOS}{non-line-of-sight}
\acrodef{GDOP}{geometric dilution of precision}
\acrodef{GPS}{Global Positioning System}
\acrodef{FIM}{Fisher information matrix}
\acrodef{PEB}{position error bound}
\acrodef{SPEB}{squared position error bound}
\acrodef{TOA}{time-of-arrival}
\acrodef{TOF}{time-of-flight}
\acrodef{WSN}{wireless sensor network}
\acrodef{MAC}{medium access control}
\acrodef{RSS}{received signal strength}
\acrodef{WAF}{wall attenuation factor}
\acrodef{TDOA}{time difference-of-arrival}
\acrodef{RF}{radiofrequency}
\acrodef{RTT}{round-trip time}
\acrodef{AOA}{angle-of-arrival}
\acrodef{MF}{matched filter}
\acrodef{ED}{energy detector}
\acrodef{ML}{maximum likelihood}
\acrodef{MSE}{mean-square error}
\acrodef{RMSE}{root-mean-square error}
\acrodef{LEO}{localization error outage}
\acrodef{ppm}{part-per-million}
\acrodef{ACK}{acknowledge}
\acrodef{UWB}{Ultrawide bandwidth}
\acrodef{TNR}{threshold-to-noise ratio}
\acrodef{LS}{least squares}
\acrodef{IR-UWB}{impulse radio UWB}
\acrodef{FCC}{Federal Communications Commission}
\acrodef{TH}{time-hopping}
\acrodef{PPM}{pulse position modulation}
\acrodef{MUI}{multi-user interference}
\acrodef{PDP}{power delay profile}
\acrodef{BPZF}{band-pass zonal filter}
\acrodef{SIR}{signal-to-interference ratio}
\acrodef{SINR}{signal-to-interference-plus-noise ratio}
\acrodef{RFID}{radio frequency identification}
\acrodef{WPAN}{wireless personal area network}
\acrodef{WWB}{Weiss-Weinstein bound}
\acrodef{DP}{direct path}
\acrodef{MF}{matched filter}
\acrodef{MMSE}{minimum-mean-square-error}
\acrodef{SBS}{serial backward search}
\acrodef{SBSMC}{serial backward search for multiple clusters}
\acrodef{NBI}{narrowband interference}
\acrodef{WBI}{wideband interference}
\acrodef{INR}{interference-to-noise ratio}
\acrodef{CR}{channel response}
\acrodef{CIR}{channel impulse response}
\acrodef{CR}{channel  response}
\acrodef{RADAR}{radar}
\acrodef{MUR}{Multistatic radar}
\acrodef{JBSF}{jump back and search forward}
\acrodef{HDSA}{high-definition situation-aware}
\acrodef{RRC}{root raised cosine}
\acrodef{ST}{simple thresholding}
\acrodef{BTB}{Bellini-Tartara bound}
\acrodef{P-Max}{$P$-Max}  %suggestion, use with \acl{P-Max}
\acrodef{MIMO}{multiple-input multiple-output}
\acrodef{MAP}{maximum a posteriori}
\acrodef{FG}{factor graph}
\acrodef{OP}{outage probability}
\acrodef{WED}{wall extra delay}
\acrodef{RMS}{root mean square}
\acrodef{SPAWN}{sum-product algorithm over a wireless network}
\acrodef{MDD}{minimum distance distribution}
\acrodef{MAP}{maximum a posteriori probability}
\acrodef{SAP}{small cell access point}
\acrodef{UE}{user equipment}
\acrodef{MBS}{macro cell base station}
\acrodef{UER}{\ac{UE} Relay}
\acrodef{D2D}{device-to-device}
\acrodef{MBS}{macro base station}
\acrodef{CSI}{channel state information}
\acrodef{OGR}{outage guard region}
\acrodef{FUR}{feasible UER region}
\acrodef{EHR}{energy harvesting region}
\acrodef{EH}{energy harvesting}
\acrodef{D2D-EHSN}{D2D communication provided \ac{EH} small cell network}
\acrodef{D2D-EHHN}{D2D communication provided \ac{EH} heterogeneous network}
\acrodef{3GPP}{3rd Generation Partnership Project}
\acrodef{BS}{base station}
\acrodef{DF}{decode and forward}
\acrodef{CCDF}{complementary cumulative distribution function}
\acrodef{ZF}{zero forcing}
\acrodef{RZF}{regularized zero forcing}
\acrodef{WLLN}{weak law of large number}
\acrodef{SLLN}{strong law of large numbers}
\acrodef{TDD}{Time-division duplex}
\acrodef{EE}{energy efficiency} 
\acrodef{HetNet}{heterogeneous network} 
\acrodef{SCP}{Single Cell Processing}
\acrodef{CBF}{Coordinated Beamforming}
\usepackage{color}
\usepackage{dsfont}
\usepackage{bbm}

\DeclareMathAlphabet{\mathsf}{OML}{cmbr}{m}{it}

\newtheorem{theorem}{\bf Theorem}
\newtheorem{lemma}{\bf Lemma}
\newtheorem{corollary}{\bf Corollary}

\newtheorem{remark}{\bf Remark}

\newtheorem{assumption}{\bf Assumption}

\newcommand{\bd}{\begin{description}}
\newcommand{\ed}{\end{description}}
\newcommand{\be}{\begin{enumerate}}
\newcommand{\ee}{\end{enumerate}}
\newcommand{\bi}{\begin{itemize}}
\newcommand{\ei}{\end{itemize}}
\newcommand{\bl}{\begin{list}}
\newcommand{\el}{\end{list}}
\newcommand{\bt}{\begin{tabbing}}
\newcommand{\et}{\end{tabbing}}

\newcommand{\paperTitle}{Rethinking Federated Learning Over the Air: \\
The Blessing of Scaling Up}

\begin{document}

{
\title{\paperTitle}

\author{

	\vspace{0.2cm}
	    Jiaqi~Zhu, 
        Bikramjit Das,
             Yong Xie, \textit{Member, IEEE},
        Nikolaos Pappas, \textit{Senior Member, IEEE},           
        and 
        Howard~H.~Yang, \textit{Member, IEEE}
       
		% $^\dagger$ \textit{ZJU-UIUC Institute, Zhejiang University, China }\\
          % $^\mathsection$ \textit{Department of Computer and Information Science, Linköping University, Sweden}
          % \\
          
% % \thanks{The work of J. Zhu and H.~H.~Yang was supported in part by the National Natural Science Foundation of China under Grant 62201504, in part by the Zhejiang Provincial Natural Science Foundation of China under Grant LGJ22F010001, and in part by the Zhejiang – Singapore Innovation and AI Joint Research Lab. The work of N. Pappas has been supported by the Swedish Research Council (VR), ELLIIT, the European Union (ETHER, 101096526), and the European Union's Horizon Europe research and innovation programme under the Marie Skłodowska-Curie Grant Agreement No. 101131481 (SOVEREIGN).} 
        \thanks{ J. Zhu and H.~H.~Yang are with the Zhejiang University/University of Illinois Urbana-Champaign Institute, Zhejiang University, Haining 314400, China; H.~H.~Yang is also with the Department of Electrical and Computer Engineering at the University of Illinois Urbana-Champaign, IL, USA (e-mail: haoyang@intl.zju.edu.cn).}	

        \thanks{ B.~Das is with the Engineering Systems and Design, Singapore University of Technology and Design, Singapore 487372 (e-mail: bikram@sutd.edu.sg).}

        \thanks{ Y.~Xie is with the School of Computer Science, Nanjing University of Posts and Telecommunications, Nanjing 210000, China (e-mail: yongxie@njupt.edu.cn).}         

         \thanks{ N. Pappas is with the Department of Computer and Information Science, Linköping University, Linköping 58183, Sweden (e-mail: nikolaos@liu.se).}

}
\maketitle
\acresetall
\thispagestyle{empty}
\begin{abstract}
Federated learning facilitates collaborative model training across multiple clients while preserving data privacy. However, its performance is often constrained by limited communication resources, particularly in systems supporting a large number of clients. To address this challenge, integrating over-the-air computations into the training process has emerged as a promising solution to alleviate communication bottlenecks.
The system significantly increases the number of clients it can support in each communication round by transmitting intermediate parameters via analog signals rather than digital ones. This improvement, however, comes at the cost of channel-induced distortions, such as fading and noise, which affect the aggregated global parameters.
To elucidate these effects, this paper develops a theoretical framework to analyze the performance of over-the-air federated learning in large-scale client scenarios. Our analysis reveals three key advantages of scaling up the number of participating clients:
(1) Enhanced Privacy: The mutual information between a client’s local gradient and the server’s aggregated gradient diminishes, effectively reducing privacy leakage.
(2) Mitigation of Channel Fading: The channel hardening effect eliminates the impact of small-scale fading in the noisy global gradient.
(3) Improved Convergence: Reduced thermal noise and gradient estimation errors benefit the convergence rate.
These findings solidify over-the-air model training as a viable approach for federated learning in networks with a large number of clients. The theoretical insights are further substantiated through extensive experimental evaluations.
\end{abstract}

\acresetall

%%%%%%%%%%%%%%%%%%%%%%%%%%%%%%%%%%%%%%%%%%%%%%%%%%%%
\section{Introduction}\label{sec:intro} 

Federated learning (FL) is an emerging distributed machine learning paradigm that enables collaborative training of a global model across multiple clients without exposing their private local data to a central server \cite{mcmahan17communication, li20federated}. Unlike traditional centralized approaches, FL decentralizes the learning process by offloading the training to the clients, exploiting end-user devices' processing and storage capabilities. Typically, each round of model training consists of three stages: local update and parameter uploading at the clients, parameter aggregation and global model update at the edge server, and global model broadcasting from the server to clients for the next round of training. While promoting data privacy by keeping local data in-device, the frequent exchange of parameters between the edge server and clients incurs significant communication overhead, which becomes a paramount performance bottleneck for FL, especially when the number of participating clients becomes large \cite{niknam20federated}.

In response, a line of recent studies \cite{zhu19broadband, yang20federated, sery20analog, amiri20federated} suggested integrating over-the-air computations \cite{nazer07computation} into the FL model training procedure, leveraging the superposition nature of multiple access channels for fast and scalable parameter aggregation.
Specifically, under this scheme, each client modulates its intermediate parameters (such as local gradient) onto a set of common waveforms (which are orthogonal to each other) and simultaneously transmits the analog signal to the edge server.
Subsequently, the edge server can extract an automatically aggregated gradient from the received signal to update the global model and feed it back to the clients for further local training.
As opposed to digital communication-based parameter uploading, analog transmission prevents the linear increase in spectral resource consumption as the number of participating entities grows.
Moreover, increasing the number of clients can improve energy efficiency, thereby reducing the transmit power \cite{sery20analog}.  
In addition, analog over-the-air computing bypasses the encoding (resp. decoding) and modulation (resp. demodulation) processes typically required before aggregation, which significantly reduces the access latency \cite{zhu19broadband,yang20federated,zhao24model}.
As a result, adopting over-the-air computations enables the implementation of large-scale edge learning systems with low-cost communication modules \cite{yang21revisiting,yang24unleashing}.

However, these are achieved at the price of introducing additional distortions into the received signal. Indeed, analog signals transmitted over the spectrum are inevitably corrupted by channel fading and thermal noise, rendering the aggregated gradient a noisy version of the desired one \cite{chen23edge}. Directly using this noisy gradient for model update results in unstable training performance or even impedes the efficacy of model convergence, thus calling for developing optimization strategies to mitigate aggregation distortion. To that end, several works \cite{liu20over, cao20optimized, cao21optimized, guo22joint} suggested that every client estimate its instantaneous channel quality before each global transmission, such that they can adequately control their power to counteract wireless channel defects in the aggregated global signal.
As a result, it reduces fluctuations in the training process and achieves stable convergence. In addition, \cite{jing22federated} and \cite{yu23optimal} proposed using the statistical channel state information (CSI) to adaptively adjust each client's transmit power to enhance the learning performance. 
When the base station is equipped with a large number of antennas, \cite{amiri21blind} and \cite{wei23random} further reduced the CSI estimation cost by exploiting the channel hardening effect in massive multiple-input and multiple-output (MIMO) systems.
Unfortunately, when a large number of clients are present in the system (which is the appropriate case for adopting over-the-air computations in FL training), accurately estimating the instant CSI becomes exceptionally costly—moreover, frequent CSI exchanges between clients and the edge server result in considerable communication overheads. Besides, errors associated with channel estimation are inevitable in practice, whereby imperfect CSI would lead to imperfect channel inversion and inaccurate truncation decisions, affecting the training convergence. This naturally leads to a crucial question: \textit{Is instant channel estimation and the subsequent power control necessary for federated model training over the air in large-scale networks?} 
The answer from this paper is no.
% We conclude in this paper that the answer to the previous case is no.

On the other hand, while channel distortions are commonly regarded as detrimental to training efficiency, it is also consensual that they contribute positively to enhancing end-user privacy during model training. Although privacy protection is emphasized as a salient feature of FL, as it eliminates the need for clients to directly share their raw local data by instead exchanging model/gradient updates, various studies have shown that this alone is not sufficient to ensure privacy \cite{zhu19deep, geiping20inverting}. For instance, private data information can still be inferred from individual updates through membership inference attack \cite{shokri17membership} or model inversion attack \cite{geiping20inverting}. In contrast, aggregating intermediate parameters over the air ensures potential eavesdroppers can only access the aggregated updates, thereby providing privacy protection to the participating clients. 
A few existing studies have identified this aspect \cite{seif20wireless, koda20differentially, liu20privacy, elgabli21harnessing, krouka22communication}. 
Specifically, \cite{seif20wireless} demonstrated that the wireless channel provides a dual benefit of bandwidth efficiency together with strong local differential privacy (DP) guarantees, with the privacy level per user scaling as $\mathcal{O}(1/\sqrt{N})$. Similarly, \cite{koda20differentially} proposed harnessing the inherent receiver noise to preserve DP against inference attacks, highlighting the number of clients as a key factor in maintaining a higher privacy level. And \cite{liu20privacy} showed that privacy can be obtained ``for free'' from the channel noise when the privacy constraint level is below a certain threshold. In \cite{elgabli21harnessing} and \cite{krouka22communication}, privacy concerns are addressed by incorporating channel perturbations into the optimization problem and employing second-order methods that avoid explicitly transmitting the Hessian matrix or gradient to the edge server, thereby preserving data privacy against model inversion attacks. Nevertheless, how much privacy is guaranteed or how much information the aggregated model updates leak about a single client's local dataset remains unclear. The present paper develops an information-theoretic metric to measure privacy leakage in the context of over-the-air model aggregation, affirming that the more clients present in the system, the less privacy leakage from each individual.

In brief, although adopting over-the-air computations in model aggregation is originally regarded as an alternative to circumvent the communication bottleneck when the system needs to support a large number of clients, the key finding in this paper reveals that the reverse is also true, i.e., \textit{the presence of massive amount of clients enhances the performance of federated learning over the air.}
% The technique and its application scenario form an ideal match, and the two shall end up in a happy marriage.\textcolor{red}{maybe too much as a statement :)}

The technical contributions are summarized below.
\begin{itemize}
    \item We quantify the privacy leakage of analog over-the-air model aggregation by deriving an analytical expression for the mutual information (MI) between each client's locally possessed gradient and the globally aggregated one. The analysis demonstrates a two-fold benefit of model training over the air: On the one hand, distortions stemming channel fading and thermal noise reduce privacy leakage; on the other hand, with a large number of clients in the system, each participant's MI approaches zero.      

    \item We establish a concentration inequality for the noisy gradient obtained from over-the-air model aggregation, which unveils a channel-hardening effect in the global averaging stage. 
    Subsequently, we derive analytical expressions for the convergence rate for both convex and non-convex loss functions by taking into account data heterogeneity. 
    Grounded on the prior literature, we emphasize and quantify the role of the number of clients in the convergence in a comprehensive way.
    The result shows that as the number of clients increases, the impairments from communication and estimation noise vanish, thus expediting a stable convergence process. We also derive the convergence rate under power control, accounting for the effect of imperfect CSI. Our result implies that power control is not essential in large-scale over-the-air FL systems.
    
    \item We conduct extensive simulations to verify the theoretical findings. Specifically, we conduct learning tasks with the convolutional neural network (CNN) and ResNet-18 on the EMNIST and CIFAR-10 datasets under different system configurations. The experiments confirm that increasing the number of clients improves privacy protection, training efficiency, and system robustness, validating the benefits of scaling up the system.
\end{itemize}

\textit{Notations}: Throughout the paper, bold lowercase letters represent column vectors. Given a vector $\boldsymbol{w}$, $\boldsymbol{w}^{\top}$ denotes its transpose and $\|\boldsymbol{w}\|$ denotes its L-2 norm. The identity matrix of dimension $d\times d$ is denoted by $\boldsymbol{I}_d$. For any positive integer $i$, $[i]$ denotes the set of integers $\{1,2,...,i\}$, and $\boldsymbol{w}_{(i)}$ denotes the $i$-th entry of the vector $\boldsymbol{w}$.

\begin{figure*}[t!]
    \centering
    \includegraphics[width=0.85\textwidth]{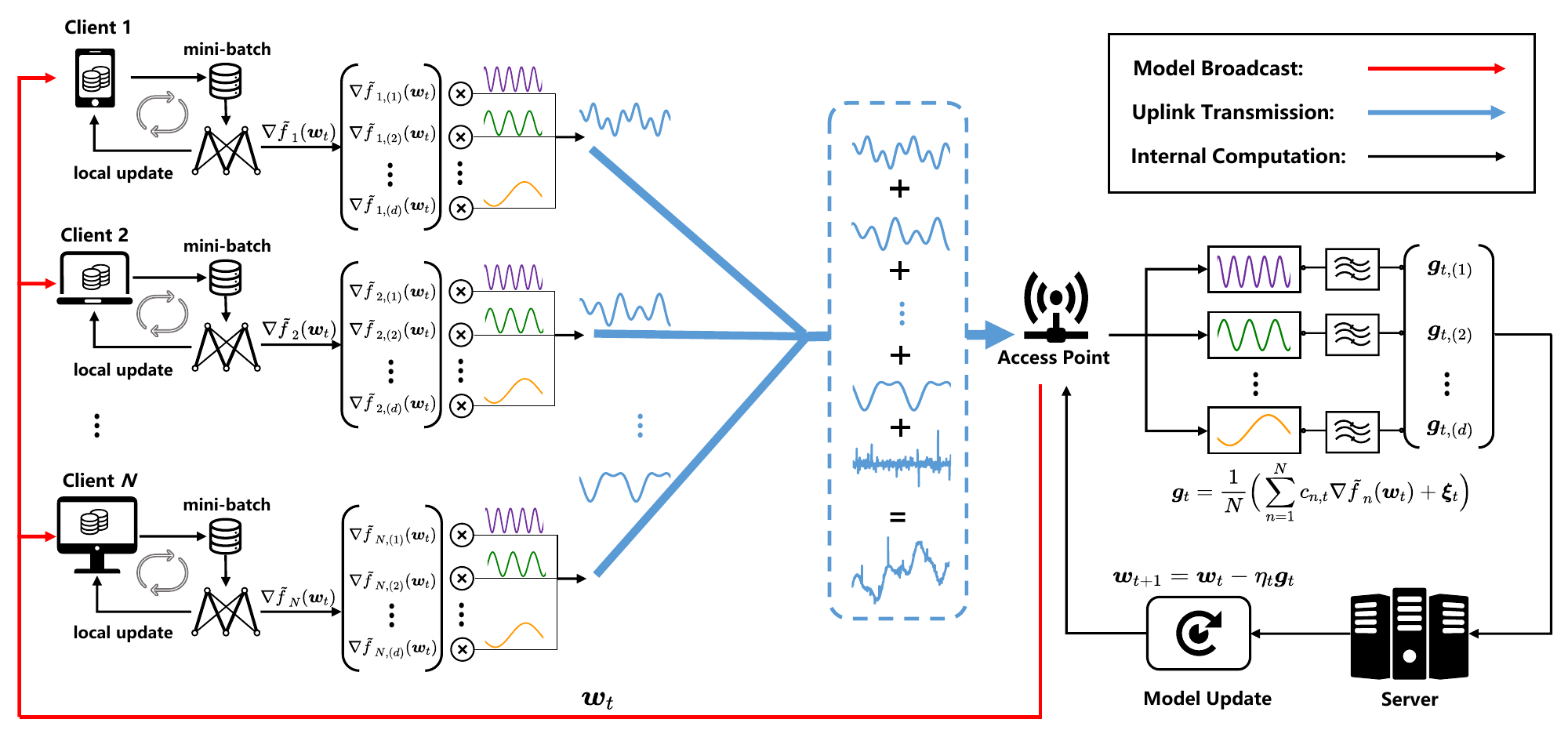} 
    \caption{An overview of the edge learning system. 
    The following steps are repeated until the model converges: (1) each client uses mini-batches sampled from its local dataset to perform several local updates and uploads the locally accumulated gradient to the server via analog transmissions; (2) the server extracts an automatically aggregated but noisy global gradient from the received radio signal, and uses it to update the global model; (3) the updated model is broadcast to all the clients for a new round of local training. 
    }
\label{fig:system_model}
\end{figure*}

%%%%%%%%%%%%%%%%%%%%%%%%%%%%%%%%%%%%%%%%%%%%%%%%%%%%
\section{System Model}\label{sec:sysmod}
%%------------------------------------------------%%
\subsection{Setting}

We consider the federated edge learning system depicted in Fig.~\ref{fig:system_model}, which comprises an edge server and $N$ clients, where $N \gg 1$. Every client $n$ holds a local dataset $\mathcal{D}_{n} = \{( \boldsymbol{x}_{i}, y_{i} ) \}_{i=1}^{m_n}$ where $\boldsymbol{x}_{i} \in \mathbb{R}^{d}$ and $y_{i} \in \mathbb{R}$ represent the data sample and the corresponding response, respectively. We assume the local datasets are statistically independent but not necessarily identically distributed. For simplicity, we assume the clients have equal-sized datasets, i.e., $m_n=M$, $\forall n \in [N]$.

The goal of all the entities in this system is to collaboratively train a statistical model from the clients' data samples without sacrificing their data privacy. More precisely, they need to jointly minimize the following objective function:
\begin{equation}\label{eq:Gbl_loss}
    f(\boldsymbol{w})=\frac{1}{N}\sum^{N}_{n=1}f_{n}(\boldsymbol{w})
\end{equation}
where $f_{n}(\boldsymbol{w})$ is the local empirical risk of client $n$, given by
\begin{equation}
    f_{n}(\boldsymbol{w})=\frac{1}{m_{n}}\!\sum^{m_{n}}_{i=1}\!\ell(\boldsymbol{w};\boldsymbol{x}_{i},y_{i})=\frac{1}{M}\!\sum_{i\in \mathcal{D}_{n}}\!\ell(\boldsymbol{w};i)
\end{equation}
in which $\ell(\boldsymbol{w};\boldsymbol{x}_{i},y_{i})$ quantifies the loss associated with the sample pair $(\boldsymbol{x}_{i},y_{i})$, and we simplify its notation by $\ell(\boldsymbol{w};i)$ for convenience. The optimal solution of \eqref{eq:Gbl_loss} is commonly known as the empirical risk minimizer, denoted by
\begin{equation}
    \boldsymbol{w}^{*}=\arg\min_{ \boldsymbol{w} \in \mathbb{R}^d } f(\boldsymbol{w}).
\end{equation}

%%------------------------------------------------%%
\subsection{Analog Over-the-Air Model Training}

The general procedure of federated model training under analog over-the-air computing has been detailed in \cite{yang21revisiting}. We briefly describe it in this part for completeness. At the $t$-th round of global communication, the edge server broadcasts the current global model $\boldsymbol{w}_t$ to all the clients. Subsequently, each client $n$ initializes its local model as $\boldsymbol{w}^{(0)}_{n,t}=\boldsymbol{w}_t$,
% and performs $\gamma$ local training epochs.
% At the beginning of each local epoch, client $n$ 
randomly shuffles its local dataset $\mathcal{D}_{n}$ and splits the dataset into $\lfloor\!\frac{M}{B}\!\rfloor$ mini batches, each containing $B$ data samples. Then, the client performs $E$ stochastic gradient descent steps based on the mini batches. More precisely, the local model of client $n$ at the $k$-th local iteration, $k\in [E]$, is updated as
\begin{equation}
    \boldsymbol{w}^{(k+1)}_{n,t}=\boldsymbol{w}^{(k)}_{n,t} - \eta_{l}\frac{1}{B}\!\sum^{B}_{j=1}\nabla \ell(\boldsymbol{w}^{(k)}_{n,t};\theta^{(k)}_{n,t}(j))
\end{equation}
where $\eta_{l}$ denotes the local learning rate and $\theta_{n,t}$ represents a permutation to the local dataset (of client $n$ at the $t$-th global iteration). 
Upon the completion of local training, client $n$ has its accumulated local gradient as follows:
% When local training finishes, the accumulated local gradient $\nabla \widetilde{f}_{n}( \boldsymbol{w}_t )$ can be written as
\begin{equation} 
    \nabla \widetilde{f}_{n}(\boldsymbol{w}_t)=\sum^{E-1}_{k=0}\!\frac{1}{B}\!\sum^{B}_{j=1}\!\nabla \ell(\boldsymbol{w}^{(k)}_{n,t};\theta^{(k)}_{n,t}(j)).
\end{equation}
After that, every client modulates its accumulated local gradients onto the magnitude of a set of common baseband waveforms, which are orthogonal to each other, in an entry-wise manner and simultaneously transmits the analog signal to the edge server.
We consider the clients employ power control to compensate for large-scale path loss (which varies slowly and can be accurately estimated via long-term averages of the received signal strength \cite{li06rss}) while the instantaneous channel fading remains unknown. 
Each client is subject to a maximum transmit power budget at each communication round. {\footnote{Owing to the finite power budget at the radio front end, if some entries of the upload parameter are excessively large, they need to be trimmed before being modulated to the radio signal and sent out. }}
% Each client is subject to a maximum power budget at each round $t$ and a maximum average transmit power budget over the entire training period.

Due to the superposition property of multiple access channels, the edge server can pass the received signal to a bank of matched filters, with each branch tuning to one of the waveform bases (cf. the right part of Fig.~\ref{fig:system_model}), and outputs the automatically aggregated (but distorted) gradient as follows: 
\begin{equation} \label{equ:AccmlGrdt}
    \boldsymbol{g}_t=\frac{1}{N}\Big(\sum_{n=1}^{N}c_{n,t}\nabla \widetilde{f}_{n}(\boldsymbol{w}_t)+\boldsymbol{\xi}_t\Big)
\end{equation}
in which $c_{n,t}$ represents the channel fading experienced by client $n$ and $\boldsymbol{\xi}_t \in \mathbb{R}^{d}$ is a vector comprised of the thermal noise. We assume the channel fading varies independently and identically distributed (i.i.d.) across the clients and communication rounds, and it remains constant during the transmission of local updates. 
The mean and variance of the channel fading are denoted by $\mu_c$ and $\sigma^2_c$, respectively. 
Moreover, we model the thermal noise as an additive white Gaussian noise (AWGN) with variance $\sigma^2_{z}$, i.e., $\boldsymbol{\xi}_t \sim \mathcal{N}(\boldsymbol{0}, \sigma^2_{z}\boldsymbol{I}_d)$.

Using \eqref{equ:AccmlGrdt}, the edge server updates the global model as 
\begin{equation} \label{equ:OTA_model_training_vanilla}
    \boldsymbol{w}_{t+1}=\boldsymbol{w}_t - \eta_t\boldsymbol{g}_t
\end{equation}
where $\eta_t$ is the global learning rate. The global parameter will be broadcast to all the clients for the next round of local training. Such iterations repeat until the model converges.
\begin{remark}
    \textit{
    Although each client modulates its raw gradient information directly onto the magnitude of the radio waveform bases, the analog signals have to undergo a wireless communication medium that distorts the original signals (through the effects of channel fading and thermal noise) before accumulating at the edge server. The edge server can only extract an automatically aggregated gradient from the received signal without accessing each client's information. In that respect, analog over-the-air computation constitutes a form of secure aggregation~\cite{bonawitz17SA}.    
    }
\end{remark}

%%------------------------------------------------%%
\subsection{Performance Metric}

We evaluate the system performance from the perspective of privacy leakage and training efficiency. 
Specifically, we apply mutual information (MI), a metric that quantifies the amount of common information between two entities, to assess privacy leakage~\cite{elkordy22privacyFL}. For a typical client $n$, the MI between its local gradient information and the global one in communication round $t$ is defined as
\begin{align} 
    &I^{(t)}_{N}=\max_{n\in[N]} I\left(\nabla \widetilde{f}_{n}(\boldsymbol{w}_t); \boldsymbol{g}_t\Big| \lbrace \boldsymbol{g}_{p} \rbrace_{p\in[t-1]} \right).     
\end{align}

We measure the training efficiency by the rate at which the (time-average) global gradient approaches zero with communication rounds (a.k.a. the convergence rate).
Formally, after $T$ rounds of global iterations, the convergence rate can be quantified by the following 
\begin{align}
    {R}^{(T)}_N = \frac{1}{T}\sum_{t=0}^{T-1}\mathbb{E}\left[ \|\nabla f(\boldsymbol{w}_t)\|^2 \right]
\end{align}
which is a standard convergence measure in non-convex optimization.

\begin{remark}    
    \textit{Another widely adopted metric to quantify privacy leakage is differential privacy (DP)~\cite{dwork14algorithmic, wei20federated}, which ensures that apart from the knowledge that an adversary has known, limited additional information about an individual is exposed in the released data. 
    While the worst-case privacy guarantee provided by DP is important, the chance of such scenarios occurring is relatively small.     
    Therefore, this paper employs MI to capture the average information leakage inherent in the system.
    It is noteworthy that MI exhibits certain consistency with differential privacy~\cite{wang16relation}, whereby the gap between the DP level of the MI optimal mechanism and the optimal DP level is bounded.
    And their combination, the MI-DP, can be bounded by $(\epsilon,\delta)-DP$ and $\epsilon-DP$~\cite{cuff16differential}, offering insights into privacy leakage.
    }
\end{remark}

%%%%%%%%%%%%%%%%%%%%%%%%%%%%%%%%%%%%%%%%%%%%%%%%%%%%

\section{Analysis}
This section forms the core technical contribution of the paper. We derive analytical expressions for the mutual information and convergence rate of the proposed edge learning system. Additionally, we analyze the impact of system scaling on these metrics, leveraging the derived results. The detailed proofs and mathematical derivations are presented in the Appendix for clarity and conciseness.
%%------------------------------------------------%%
\subsection{Mutual Information}
To focus on the analysis of privacy leakage, we adopt the honest-but-curious model~\cite{bonawitz17SA}, wherein the server honestly performs the required operations but may attempt to infer sensitive information about individual clients from the aggregated gradients. Consequently, we exclude scenarios involving malicious attackers or adversarial behavior.  

Since the local datasets $\{ \mathcal{D}_{n} \}_{n=1}^N$ are sampled independently, conditioned on the received global model $\boldsymbol{w}_t$, the distribution of the local stochastic gradients $\nabla \widetilde{f}_{n}(\boldsymbol{w}_t)$, $\forall n \in [N]$, are also independent of each other (but not necessarily identically distributed). 
Moreover, after certain preprocessing steps, the local gradient entries can be statistically independent of each other.{\footnote{Note that common decorrelating techniques such as random flipping \cite{ma22over} can reduce correlations between the entries of the local gradient.}}
Let $d_n^*$ denote the rank of the covariance matrix of $\nabla \widetilde{f}_{n}(\boldsymbol{w}_t)$, where $d_n^* \leq d$.
To simplify the factors under consideration, we assume that the covariance matrices of all clients' local gradients share the same rank, i.e. $d_n^*=d^*$.

% With the above statements, we can now characterize the information leakage of a single round below.
We are now ready to present the information leakage in a single communication round, which is detailed below. 
\begin{theorem}\label{theo:MI}
    \textit{
    Let $\mathcal{S}_{\boldsymbol{g}}$ denote the set of subvectors of dimension $d^*$ of $\nabla \widetilde{f}_{n}(\boldsymbol{w}_t)$ that have a non-singular covariance matrix. 
    If there exists $\bar{\boldsymbol{g}}\in \mathcal{S}_{\boldsymbol{g}}$ with $\mathbb{E} [ |\bar{\boldsymbol{g}}_{(i)}|^{4} ] < \infty$ for all $i\in [d^*]$, then $\exists C_{\bar{\boldsymbol{g}}}>0$, with which we can upper bound $I^{(t)}_{N}$ as follows:    
    \begin{align} 
        I^{(t)}_{N}\leq\frac{C_{\bar{\boldsymbol{g}}}d^*}{N - 1} + 
        \frac{1}{2}\!\sum^{d^*}_{i=1}\log{}{\!\left(\frac{\sum^{N}_{n=1}\sigma^2_{n,t,i} + \sigma^2_z}{\sum^{N-1}_{n=1}\sigma^2_{n,t,i}\!+\!\sigma^2_z}\right)}        
    \end{align}  
    where $\sigma^2_{n,t,i}$, $\forall i \in [d^*]$, represents the variance of the $i$-th entry of $c_{n,t}\nabla \widetilde{f}_{n}(\boldsymbol{w}_t)$ after preprocessing and the constant $C_{\bar{\boldsymbol{g}}}$ is associated with the finite fourth moment of preprocessed $c_{n,t}\nabla \widetilde{f}_{n}(\boldsymbol{w}_t)$.
    }
\end{theorem}
\begin{IEEEproof}
    Please refer to Appendix~A.
\end{IEEEproof}

An immediate result from the above is the MI under noiseless transmission with i.i.d. data distribution (in this case, we have $\sigma^2_{z}=0$, and $\sigma^2_{n,t,i} = \sigma^2_{t,i}$, $\forall n \in [N]$), where the MI in the $t$-th communication round can be bounded by \cite{elkordy22privacyFL}
    \begin{align} \label{equ:I_Nt_noiseless}
        I^{(t)}_{N}\leq\frac{C_{\bar{\boldsymbol{g}}}d^*}{N - 1}  + \frac{d^*}{2} \log{}{ \Big( \frac{N}{N - 1} \Big)}.
    \end{align}
By contrast, the MI bound under noisy transmission (with i.i.d. data distribution across all the clients) is
\begin{align} \label{equ:I_Nt_bnd}       
    I^{(t)}_{N}\leq\frac{C_{\bar{\boldsymbol{g}}}d^*}{N\!-\!1}\!+\!
    \frac{1}{2}\!\sum^{d^*}_{i=1}\log{}{\!\left(\frac{N \sigma^2_{t,i}\!+\!\sigma^2_z}{(N-1)\sigma^2_{t,i}\!+\!\sigma^2_z}\right)}.
\end{align}
Since $\frac{N \sigma^2_{t,i}+\sigma^2_z}{(N-1)\sigma^2_{t,i}+\sigma^2_z} < \frac{N}{N-1}$, the result in the right-hand-side of \eqref{equ:I_Nt_bnd} is smaller than that of \eqref{equ:I_Nt_noiseless}, indicating that channel distortions introduced by over-the-air computations can enhance privacy protection.
This finding also suggests that stringent power control may not always be beneficial in large-scale over-the-air FL systems (as even with ideal power control that achieves noiseless gradient transmissions, it gives rise to higher potential of information leakage).

%%------------------------------------------------%%
\subsection{Convergence Rate}
In this part, we derive the convergence rate of the considered edge learning system.
To facilitate the analysis, we make a few assumptions as follows.
\begin{assumption}[$\lambda$-strongly convexity]\label{assm:lambda-strongly convex}
    \textit{
    The functions $f_n: \mathbb{R}^{d} \rightarrow \mathbb{R}, \forall n \in [N]$, are all $\lambda$-strongly convex, i.e, for any  $\boldsymbol{w}, \boldsymbol{v} \in \mathbb{R}^{d}$, it is satisfied:
    \begin{align}
        % &f(\boldsymbol{w}) \geq f(\boldsymbol{v}) + \langle \nabla f(\boldsymbol{v}), \boldsymbol{w}- \boldsymbol{v} \rangle + \frac{\lambda}{2} \Vert \boldsymbol{w} - \boldsymbol{v}\Vert ^2, \\
        &f_n(\boldsymbol{w}) \geq f_n(\boldsymbol{v}) + \langle \nabla f_n(\boldsymbol{v}), \boldsymbol{w}- \boldsymbol{v} \rangle + \frac{\lambda}{2} \Vert \boldsymbol{w} - \boldsymbol{v}\Vert ^2
    \end{align}
    where $\lambda$ is a positive constant.        
    }
\end{assumption}

Under Assumption~1, we define the degree of data heterogeneity \cite{li19convergence} as $\Gamma=f(\boldsymbol{w^*})-\frac{1}{N}\sum f_n^*$, where $f_n^* \triangleq \min_{\boldsymbol{w}} f_n(\boldsymbol{w})$. 
This quantity measures the discrepancy between the global objective minimum and the average of the local minimum, reflecting the level of data heterogeneity.
Specifically, an increase in $\Gamma$ signals a higher degree of data heterogeneity, while $\Gamma$ takes the value zero when the clients' data distributions are i.i.d. from each other.

\begin{assumption}[$L$-smoothness]\label{assm:L-smooth}
    \textit{
    % (Lipschitz-Continuous Gradient)
    The functions $f_n: \mathbb{R}^{d} \rightarrow \mathbb{R}, \forall n \in [N]$, are all $L$-smooth, i.e, for any  $\boldsymbol{w}, \boldsymbol{v} \in \mathbb{R}^{d}$, it is satisfied:
    \begin{align}
        % &f(\boldsymbol{w}) \leq f(\boldsymbol{v}) + \langle \nabla f(\boldsymbol{v}), \boldsymbol{w}- \boldsymbol{v} \rangle + \frac{L}{2} \Vert \boldsymbol{w} - \boldsymbol{v}\Vert ^2,\\
        &f_n(\boldsymbol{w}) \leq f_n(\boldsymbol{v}) + \langle \nabla f_n(\boldsymbol{v}), \boldsymbol{w}- \boldsymbol{v} \rangle + \frac{L}{2} \Vert \boldsymbol{w} - \boldsymbol{v}\Vert ^2
    \end{align}
    where $L$ is a positive constant.
    }
\end{assumption}

\begin{assumption}[SGD Sampling Noise]\label{assm:SGD sampling noise}
    \textit{   
    For every client $n$, the stochastic gradient $\nabla \widetilde{f}_{n}(\boldsymbol{w};\mathcal{B}_{n})=\frac{1}{B}\sum_{i\in\mathcal{B}_{n}}\nabla \ell(\boldsymbol{w};i)$, calculated based on an independent mini-batch $\mathcal{B}_{n}$ containing $B$ data samples, is an unbiased estimation of $\nabla{f}_{n}(\boldsymbol{w};\mathcal{D}_{n})$ with bounded variance, i.e.,    
    \begin{align}
        &\mathbb{E} \left[ \nabla \widetilde{f}_{n}(\boldsymbol{w};\mathcal{B}_{n}) \right] = \nabla{f}_{n}(\boldsymbol{w}; \mathcal{D}_{n}),\\
        &\mathbb{E} \left[ \big\|\nabla \widetilde{f}_{n}(\boldsymbol{w};\mathcal{B}_{n})\!-\!\nabla{f}_{n}(\boldsymbol{w}; \mathcal{D}_{n})\big\|^2 \right]
        \leq \frac{\sigma^2_{s,n}}{B}.
    \end{align}    
    }
\end{assumption} 
\begin{assumption}[Gradient Bound]\label{assm:gradient bound}
    \textit{
    The expected squared norm of the stochastic gradient of functions $f_n(\boldsymbol{w})$ is bounded, i.e., for $\forall n \in [N]$, there exists a positive constant $G_n$ such that
    \begin{equation}
        \mathbb{E}[\|\nabla \widetilde{f}_n(\boldsymbol{w})\|^2] \leq G^2_n.
    \end{equation}
    }
\end{assumption}

The above assumptions are commonly adopted in the existing works \cite{li19convergence, sery21over, amiri21blind}. In particular, Assumptions 1 to 3 hold for a broad class of objective functions used in FL systems \cite{li19convergence}, while Assumption~4 is generally valid in over-the-air computation settings due to the stipulation of maximum transmit power \cite{sery21over, amiri21blind}.
Moreover, the discrepancy in $G_n$ accounts for statistical data heterogeneity among clients.

We start by investigating the effect of channel fading on the aggregated global gradient.
By virtue of the averaging operation to the globally aggregated gradient, we can establish the following concentration inequality. 

\begin{lemma} \label{lma:Hoeffding_AggGrdt}
\textit{Let $\nu>0$ be a constant, and denote by $\beta_\nu = \mathbb{P}( \vert c_{n,t} - \mu_c \vert > \nu )$. At any communication round $t$, for the $i$-th entry of the aggregated gradient, $\forall i \in [d]$, with $G = \max_{n \in [N]} (G_n)$, the following holds for all $\varepsilon \geq 0$
\begin{align} \label{equ:Hoeffding_ineq_AggGrdt}
&\mathbb{P}\!\left( \!\frac{1}{N} \!\! \sum_{n=1}^N \!\!c_{n,t} \!\nabla \widetilde{f}_{n,(i)} ( \boldsymbol{w}_t ) \!-\! \frac{1}{N} \!\! \sum_{n=1}^N \!\mu_c \!\nabla \widetilde{f}_{n,(i)} ( \boldsymbol{w}_t ) \!\geq\! \varepsilon \!+\! 2\nu \beta_\nu^N \!G \! \right)
\nonumber\\
&\leq \beta_\nu^N + \exp\bigg(\! - N \frac{\varepsilon^2}{ 2 \nu^2 G^2 } \bigg).
\end{align}
}
\end{lemma}
\begin{IEEEproof}
Please see Appendix~B.
\end{IEEEproof}

\begin{remark}
    \textit{When $N \gg 1$, the right-hand-side of \eqref{equ:Hoeffding_ineq_AggGrdt} approaches zero (by controlling $\nu$ we can ensure $\beta_\nu \ll 1$).
    This resembles the channel hardening effect \cite{hochwald04multiple}, whereby the impairment of random channel fading is averaged out in the presence of massive clients (which is the adequate regime for employing over-the-air model aggregation).   
    }
\end{remark}  

Following Lemma~\ref{lma:Hoeffding_AggGrdt}, we can tightly approximate the global gradient in \eqref{equ:AccmlGrdt} as follows:    
\begin{align}
    \boldsymbol{g}_t 
    % & = \frac{1}{N}\!\sum_{n=1}^{N}\!c_{n,t}\!\nabla \widetilde{f}_{n}(\boldsymbol{w}_t)\!+\!\frac{1}{N}\boldsymbol{\xi}_t
    % \nonumber\\
    \approx \frac{1}{N}\!\sum_{n=1}^{N} \mu_c \!\nabla \widetilde{f}_{n}(\boldsymbol{w}_t)\!+\!\frac{1}{N}\boldsymbol{\xi}_t. 
\end{align} 

% Following Lemma~\ref{lma:Hoeffding_AggGrdt}, we can tightly approximate the global gradient as follows:    
% \begin{align}
%     \boldsymbol{g}_t & = \frac{1}{N}\!\sum_{n=1}^{N}\!c_{n,t}\!\nabla \widetilde{f}_{n}(\boldsymbol{w}_t)\!+\!\frac{1}{N}\boldsymbol{\xi}_t
%     \nonumber\\
%     &\approx \frac{1}{N}\!\sum_{n=1}^{N} \mu_c \!\nabla \widetilde{f}_{n}(\boldsymbol{w}_t)\!+\!\frac{1}{N}\boldsymbol{\xi}_t. 
% \end{align} 

We are now ready to present the convergence rate.
We begin by establishing the bound for the strongly convex and smooth loss function.
The result under a fixed learning rate is given as follows.
\begin{theorem}\label{theo:ConvexConRate}
    \textit{
    Let Assumptions 1 to 3 hold; if the global and local learning rates satisfy $\eta_t = \frac{1}{\mu_c}\eta_l$ and $\eta_l \leq \frac{1}{4L}$, respectively, then model training under the employed edge learning system converges as    
    % \begin{align}\label{equ:Cnvx_CnvgRt_FixedLrnRt}        
    %     &\mathbb{E}[\|\boldsymbol{w}_t \!-\! \boldsymbol{w}^* \|^2] \!\leq\! (1 \!-\! \lambda \eta_l )^{t E} \|\boldsymbol{w}_0 \!-\! \boldsymbol{w}^* \|^2 \!+\! \frac{\eta_l}{\lambda} \bigg[6L \Gamma  \notag \\ 
    %     &\quad\!+\! \Big( \frac{1}{N} \!+\! 2(E\!-\!1) \Big) \frac{\sum^N_{n=1} \sigma^2_{s,n}}{N B} \bigg]  \!+\! \frac{\eta_l^2}{1 \!-\!(1 \!-\! \lambda \eta_l)^E} \frac{d\sigma^2_z}{\mu_c^2 N^2}.
    % \end{align}
    \begin{align}\label{equ:Cnvx_CnvgRt_FixedLrnRt}        
    &\mathbb{E}\left[ \|\boldsymbol{w}_t \!-\! \boldsymbol{w}^* \|^2 \right] \!\leq\! (1 \!-\! \lambda \eta_l )^{t E} \|\boldsymbol{w}_0 \!-\! \boldsymbol{w}^* \|^2 \!+\! \frac{ d \eta_l^2 {\sigma^2_z}/{\mu_c^2 N^2} }{1 \!-\!(1 \!-\! \lambda \eta_l)^E} \notag \\ 
    &\quad\!+\! \frac{\eta_l}{\lambda} \bigg( 6L \Gamma \!+\! \Big( \frac{1}{N} \!+\! 2(E\!-\!1) \Big) \frac{\sum^N_{n=1} \sigma^2_{s,n}}{N B} \bigg)  .
    \end{align}
    }
\end{theorem}
\begin{IEEEproof}
    Please see Appendix~C.
\end{IEEEproof}

The above result quantitatively demonstrates the effects of local SGD iterations, data heterogeneity, and channel noise on the convergence rate. 
We can see that the estimation error decays at a linear convergence rate, where the global model converges into a noisy ball around the optimum. 
The inequality also reveals that due to the influence of data heterogeneity and channel noise, increasing the number of local epochs accelerates the model convergence rate but also expands the noisy ball. 
In contrast, reducing the local step size shrinks the noisy ball, while it also slows down the convergence rate. 

Correspondingly, we can obtain the convergence rate under a decaying learning rate. 
\begin{corollary}\label{co:ConvexConRate}   
    \textit{
    Let Assumptions 1 to 3 hold; if the global and local learning rates are set as $\eta_t = \frac{1}{\mu_c}\eta_l = \frac{\eta_{0}}{1+t}$, where $\eta_{0} \leq \frac{1}{4 \mu_c L}$, then model training under the employed edge learning system  converges as
    \begin{align}        
        &\mathbb{E}\left[ \|\boldsymbol{w}_t \!-\! \boldsymbol{w}^* \|^2 \right] \leq \frac{\kappa}{1+t}
    \end{align}
    where $\kappa$ is given by 
    \begin{align}
        \kappa \!=\! \max \Bigg\{ \! \frac{\mu_c^2 \eta_0^2 \! \left( \! 6L \Gamma \!+\! \frac{ (2N(E-1)+1) \! \sum^N_{n=1} \! \sigma^2_{s,n}}{N^2 B} \!\right)}{\lambda \mu_c \eta_0 \!-\! 1}, \|\boldsymbol{w}_0 \!\!-\! \boldsymbol{w}^* \|^2 \Bigg\}.
    \end{align}
    % $\kappa \!=\! \max \{ \frac{\mu_c^2 \eta_0^2 \left[6L \Gamma \!+\! \left( \frac{1}{N} \!+\! 2(E\!-\!1) \right) \frac{\sum^N_{n=1} \sigma^2_{s,n}}{N B} \right]}{\lambda \mu_c \eta_0 \!-\! 1}, \|\boldsymbol{w}_0 \!\!-\! \boldsymbol{w}^* \|^2 \}$.    
    }
\end{corollary}
\begin{IEEEproof}
    Please see Appendix~D.
\end{IEEEproof}

Next, we establish the convergence rate under the non-convex smooth case.
It is noteworthy that in this case, we only assume the smoothness of the loss function; hence, the result is applicable to settings involving (deep) neural networks. 

\begin{theorem}\label{theo:ConRate}
    \textit{
    Let Assumptions 2 to 4 hold; if the global and local learning rates satisfy $\eta_t = \frac{1}{\mu_c}\eta_l$ and $\eta_l = \frac{1}{L}$, respectively, then model training under the employed edge learning system converges as
    \begin{align} \label{equ:CnvgRt_FixedLrnRt}
        & {R}^{(T)}_N \leq \frac{ 2L \big( f(\boldsymbol{w}_0) \!-\! f(\boldsymbol{w}^*) \big) }{TE} + \frac{d\sigma_{z}^2}{\mu_c^2 N^2E} + \frac{\sum^N_{n=1}\sigma^2_{s,n}}{N^2 BE}
        \nonumber \\
        &\qquad \qquad \qquad \qquad \qquad + \frac{(E\!-\!1)(2E\!+\!5)\sum^N_{n=1}\!G_n^2}{6N}.
    \end{align}
    }
\end{theorem}
\begin{IEEEproof}
    Please see Appendix~E.
\end{IEEEproof}

Similarly, we also derive the convergence rate under a decaying learning rate. 
\begin{corollary}\label{co:ConRate}
    \textit{
    Let Assumptions 2 to 4 hold; if the global and local learning rates are set as $\eta_t = \frac{1}{\mu_c}\eta_l = \frac{\eta_{0}}{1+t}$, where $\eta_{0}<\frac{1}{\mu_c EL}$, then model training under the employed edge learning system converges as
    \begin{align}        
        &\min_{t=0,1,\cdots,T-1} \!\!\!\!\!\! \mathbb{E} \left[ \| \nabla f(\boldsymbol{w}_t) \|^2 \right] \leq \frac{2\big( f(\boldsymbol{w}_0) \!-\! f(\boldsymbol{w}^*) \big)}{\mu_c \eta_0 E \log_{}{T}} \!+\! \frac{L \mu_c\eta_{0}\pi^2}{3 \log_{}{\!T}}\nonumber \\
        &\!\times \!\!\bigg(\!\frac{d\sigma_{z}^2}{\mu_c^2 N^2 E}\!+\!\frac{\!\sum^N_{n=1} \!\sigma^2_{s,n} }{ N^2 \!B}   \!+\! \frac{L^2 \eta_0^2 \mu_c^2 \pi^2 E(E\!\!-\!\!1)(2E\!\!-\!\!1)}{90N}\!\!\sum^N_{n=1}\!G_n^2 \!\!\bigg)\!.
    \end{align}
    }    
\end{corollary}
\begin{IEEEproof}
    Please see Appendix~F.
\end{IEEEproof}

Several remarks are in order.
% \begin{remark}
%     \textit{
%     As over-the-air model training undergoes SGD and channel perturbations, it can escape saddle points and reach local minima~\cite{jin21nonconvex}, which could be globally optimal in a broad class of non-convex optimization problems.
%     Moreover, an appropriate level of noise reduces the likelihood of getting stuck in poor local minima, facilitating better convergence in the training process.
%     }
% \end{remark}
\begin{remark}
   \textit{
    Under a fixed learning rate, the global model ultimately lands within a noisy ball around the optimality, at the rate of $\mathcal{O}(\frac{1}{T})$. In contrast, by adopting a decaying learning rate, the model can arrive at the minima, at the cost of a slower convergence rate on the order of $\mathcal{O}( \frac{1}{\log(T)} )$.
    }
\end{remark}

\begin{remark}
    \textit{
    If we denote by local epoch $E$ as a portion of the mini-batch counts, i.e., $E = \tau \frac{M}{B}$, we can rewrite the error bound as ${R}^{(T)}_N \leq \frac{ 2L \big( f(\boldsymbol{w}_0) - f(\boldsymbol{w}^*) \big) }{ \tau M T } B + \frac{d\sigma_{z}^2}{ \tau M \mu_c^2 N^2} B + \frac{\sum^N_{n=1}\sigma^2_{s,n}}{\tau M N^2 }+\frac{ \big( \frac{ \tau M }{ B } - 1 \big) \big( 2\frac{ \tau M }{ B } + 5 \big)\sum^N_{n=1}\!G_n^2}{6N}$.
    This implies that properly tuning the batch size helps balance computational efficiency and model accuracy, reducing residual errors during training.  
    }    
\end{remark}

\subsection{Effects of Scaling Up}
The analysis derived above enables us to explore the effects of system scale-up, i.e., with an increase in the client number $N$, on the model training performance.  
Below, we demonstrate three notable benefits conferred by system scaling up.

\subsubsection{Enhancing Privacy Protection}
When $N \rightarrow \infty$, Theorem~\ref{theo:MI} yields 
\begin{align}
    I^{(t)}_N \leq \frac{C_{\bar{\boldsymbol{g}}}d^*}{N\!-\!1} + \frac{1}{2} \sum^{d^*}_{i=1} \frac{\sigma^2_{n,t,i}}{\sum^{N-1}_{n=1}\sigma^2_{n,t,i}\!+\!\sigma^2_z} \sim \mathcal{O} \Big( \frac{1}{N} \Big).
    % I^{(t)}_N \leq \frac{C_{\bar{\boldsymbol{g}}}d^*}{N\!-\!1} + \frac{1}{2} \sum^{d^*}_{i=1} \frac{\sigma^2_{n,t,i}}{\sum^{N-1}_{n=1}\sigma^2_{n,t,i}} \sim \mathcal{O} \Big( \frac{1}{N} \Big).
\end{align}
This indicates that the privacy leakage of the noisy aggregated gradient decays at the rate of $\mathcal{O}(1/N)$.
In other words, a large number of participants facilitates each individual to \textit{hide its information in the crowd}, thus enhancing privacy protection. 

\subsubsection{Mitigating Channel Impairments} 
Based on Lemma~\ref{lma:Hoeffding_AggGrdt}, we note that the difference between the desired global gradient $\frac{1}{N} \sum_{n=1}^N \nabla f_n ( \boldsymbol{w}_t )$ and its perturbed version $\frac{1}{N} \sum_{n=1}^N \frac{ c_{n,t} }{ \mu_c } \nabla f_n ( \boldsymbol{w}_t )$ decreases at the rate of $\mathcal{O}( \exp( - N ) )$, namely, the channel hardening effect (quickly) becomes evident when the number of participating clients increases.
% Such a phenomenon implies the efforts exerted in channel estimation and power control for combating the fluctuations incurred by channel fading are, in essence, unnecessary. 
Notably, while \cite{amiri21blind} shows that the fading distortions in over-the-air federated learning diminish as the number of antennas approaches infinity, our analysis unveils that with a large number of clients present, the averaging automatically ignites channel hardening, which alleviates channel impairments in over-the-air computations. 
% Similar to the results in \cite{amiri21blind}, where the fading distortion diminishes as the number of antennas approaches infinity, the fading MAC boils down to a deterministic channel with identical gains from all the clients as the number of participating clients approaches infinity.
    
\subsubsection{Improving Training Efficiency}
In addition to the channel hardening effect, a large number of participating clients also reduces estimation and communication noise, which, in turn, improves training efficiency. To be more concrete, let us take $E=1$ as an example (also known as the FedSGD). 
For the non-convex smooth case, the estimation error at global iteration $T$ can be bounded by the following
\begin{align} \label{equ:SpclCase_CvgcRt}
    R^{(T)}_N \leq \frac{ 2L \big( f(\boldsymbol{w}_0) \!-\! f(\boldsymbol{w}^*) \big) }{ T } + \frac{d\sigma_{z}^2}{ \mu_c^2 N^2 } + \frac{ \sum^N_{n=1} \sigma^2_{s,n} }{ B N^2 }.
\end{align}
This inequality indicates that increasing $N$ reduces the impact of thermal noise and, more importantly, decreases the estimation noise stemming from SGD updates, where the noise comes from using mini-batches to approximate the full gradient.
As $N$ grows, the term $\frac{1}{N^2}$ dominates, substantially decreasing the sampling noise. 
This reduction in sampling noise mitigates the effects of data heterogeneity, smoothing out the variations between clients' local gradients and promoting a more stable convergence process.
Indeed, with $N \rightarrow \infty$, the residual error (i.e., the second and third terms on the right-hand-side of \eqref{equ:SpclCase_CvgcRt}) vanishes, implying that the stochastic gradient obtained from an analog, noisy transmission would be equivalent to a global gradient achieved under a noiseless transmission condition. 

While conventional results often suggest a trade-off between privacy and training efficiency, our analytical derivations demonstrate that both metrics can be jointly enhanced as the number of clients in the system increases. In the next section, we validate these theoretical findings through simulations. Before that, we further investigate the necessity of power control in the setting of over-the-air federated learning. 

\subsection{Is Power Control Necessary?}
A plethora of prior research \cite{liu20over, cao20optimized, cao21optimized, guo22joint} have explored various power control strategies (based on instant CSI) in the context of over-the-air federated learning, aiming to reduce channel distortions in the aggregated gradient and hence enhance system performance. We argue that this is not necessarily true.

From the perspective of privacy leakage, the knowledge of the channel state increases the capacity of the fading channel \cite{thomas06elements}. As a result, perfect CSI at the transmitter would increase the mutual information between the local gradient and the global one, leading to a higher risk of privacy leakage.

From the perspective of training efficiency, power control may cause (unnecessary) client dropout, which degrades performance. More concretely, the clients experiencing poor channel conditions cannot engage in the communication round because of transmit power constraints. Additionally, the availability of perfect CSI at the transmitter, assumed by most existing works, is difficult to obtain in practice. In that respect, while the overhead of channel estimation increases with $N$, the CSI imperfection further disrupts model aggregation by causing inaccuracies in truncation decisions and channel inversion. Below, we derive the convergence rate of the edge learning system considered under power control to demonstrate these effects analytically. 

Specifically, we use the bounded channel estimation error model \cite{zhu20one} to characterize the imperfect CSI and represent the estimated CSI as $\widetilde{c}_{n,t} = c_{n,t} + \Delta_{n,t}$, where $c_{n,t}$ and $\Delta_{n,t}$ denote the actual channel gain and the estimation error at the $t$-th communication round of client $n$, respectively. 
We assume that the estimation error is bounded by $|\Delta_{n,t}| \leq \Delta_{\max} \ll c_{th}$, with a zero mean and a variance $\sigma^2_{\Delta}$. We apply the widely used truncated channel inversion scheme \cite{zhu19broadband, cao20optimized, sery21over} with a cutoff threshold $c_{th}$. 
Note that while this may be a suboptimal solution, it is a practical solution for large-scale systems and provides explicit illustration of the impact of power control on system performance.
Consequently, we set the transmit coefficient of client $n$ as
\begin{align}
    b_{n,t}=\begin{cases}
        \frac{\sqrt{\gamma_t}}{\widetilde{c}_{n,t}}, & \text{if} \quad \!\!\widetilde{c}_{n,t} \geq c_{th}, \\
        0, & \text{otherwise}
    \end{cases}
\end{align}
where $\gamma_t$ is the scalar factor for power constraint.   
Correspondingly, the aggregated gradient at the server takes the following form
\begin{align}
    \boldsymbol{g}^{P}_t=\frac{1}{|\mathcal{S}_t|}\Big(\sum_{n \in \mathcal{S}_t}\frac{c_{n,t}}{\widetilde{c}_{n,t}}\nabla \widetilde{f}_{n}(\boldsymbol{w}_t)+\frac{\boldsymbol{\xi}_t}{\sqrt{\gamma_t}}\Big)
\end{align}
where $\mathcal{S}_t$ denotes the set of clients with the estimated channel satisfying $\widetilde{c}_{n,t} \geq c_{th}$.
Then, we have the convergence rate given by the following. 
\begin{theorem}\label{theo:PowerControl}
    \textit{
    Let Assumptions 2 to 4 hold; if the global and local learning rates satisfy $\eta_t = \eta_l = \frac{1}{L}$, then model training under the employed edge learning system converges as
    \begin{align} \label{equ:CvgtRt_PwrCtrl}
    {R}^{(T)}_N \leq & \frac{ 2L \big( f(\boldsymbol{w}_0) \!-\! f(\boldsymbol{w}^*) \big) }{TE} + \frac{d\sigma_{z}^2}{S^2 \gamma_t E} + \frac{2\sum^N_{n=1}\sigma^2_{s,n}}{N^2 BE} 
        \nonumber \\
    % & +\!\! \left(\!\frac{2(E\!-\!1)(E\!+\!1)}{3} \!+\! \frac{2(N\!-\!S)}{S(N\!-\!1)} \!+\! \frac{E \sigma^2_{\Delta}}{S(c_{th}\!-\!\Delta_{\max})^2}\!\!\right)\! \frac{\sum^N_{n=1}\!G_n^2}{N}
    & +\!\frac{2(E\!-\!1)(E\!+\!1)}{3N}\!\!\sum^N_{n=1}\!G_n^2 \!+\! \frac{2(N\!-\!S)}{SN(N\!-\!1)}\!\!\sum^N_{n=1}\!G_n^2
    \nonumber \\
    & +\! \frac{E \sigma^2_{\Delta} \sum^N_{n=1}\!G_n^2}{SN(c_{th}\!-\!\Delta_{\max})^2}
    \end{align}
    where $S = N \mathbb{P}(c_{n,t}>c_{th})$.
    % $D_1$ and $D_2$ represent the additional distortion introduced by clients drop-out and imperfect CSI, respectively.
    }    
\end{theorem}
% \begin{IEEEproof}
%     Please see Appendix~G.
% \end{IEEEproof}

The last two terms on the right-hand side of \eqref{equ:CvgtRt_PwrCtrl} represent the distortion introduced by client dropout and CSI estimation error. 
Compared to \eqref{equ:CnvgRt_FixedLrnRt}, this indicates that we can circumvent the challenge of channel noise reduction by simply scaling up the system, which avoids the complexities and limitations associated with power control. 

It is important to note that we are not arguing that channel estimation is irrelevant in over-the-air FL systems. 
Instead, we emphasize that power control, which depends on instant perfect CSI, becomes unnecessary in large-scale systems. 
Nevertheless, channel estimation remains essential to compensate for the phase distortions introduced by wireless channels. 
Notably, phase correction is only needed to ensure positive channel gains at the edge server, and a correction with an error less than $\pi /2$ suffices to achieve positive gains at the receiver, meaning only partial knowledge of the channel phase is required.
To that end, over-the-air FL can be efficiently implemented on a scale.

%%------------------------------------------------%%
%               Section: Experiment 
%%------------------------------------------------%%
\section{Experimental Results}

This section evaluates the performance of the proposed framework across different system scales. We begin by describing the experimental setup, followed by a comprehensive analysis of the system’s performance from multiple perspectives, including privacy leakage quantified through mutual information, visualization of the channel hardening effect, and training efficiency. Additionally, we examine the influence of data heterogeneity and assess the effectiveness of power control. Finally, we explore the ancillary benefits of scaling up the system, emphasizing the enhanced resilience of large-scale systems against malicious attacks and their robustness in employing second-order optimization methods.

\subsection{Setup}

\subsubsection{Dataset and Models}
We evaluate the performance of the considered edge learning system by experiments on the CIFAR-10 \cite{krizhevsky09learning} and EMNIST \cite{cohen17emnist} dataset using neural network model architectures including the ResNet-18 \cite{he16deepresidual} and a Convolutional Neural Network (CNN) that consists of two convolutional layers (each followed by a max pooling layer and ReLU activation) and three fully-connected layers, respectively.
The CIFAR-10 dataset comprises 60,000 images across 10 classes, divided into a training set of 50,000 images and a test set of 10,000 images.
The EMNIST letters dataset contains 145,600 data samples belonging to 26 categories, with 124,800 images for training and 20,800 images for testing.

\subsubsection{System Configuration}
Unless otherwise stated, we model the channel gain using Rayleigh fading with an average of $\mu_c=1$.
The training set is distributed in a non-i.i.d. manner across a total of $N=100$ clients,  with variations in both class distributions and local dataset sizes. Specifically, the non-i.i.d. data partitions are implemented via the widely adopted symmetric Dirichlet distribution \cite{hsu19measuring}, in which the level of heterogeneity across clients is controlled by the coefficient $Dir$, set to 0.1 here. To assess the impact of the number of clients on the considered edge learning system, we vary the number of clients $n$ by selecting the first $n$ clients as participants accordingly. We further consider the same learning rate $\eta= 0.03$ for both local and global training (since $\mu_c=1$), local epoch $E = \lfloor\!\frac{M}{B}\!\rfloor$, and local batch size $B=50$.

All experiments are implemented with Pytorch on NVIDIA RTX 3090 GPU.

\begin{figure}[t!]
    \centering
    \includegraphics[width=0.85\columnwidth]
    {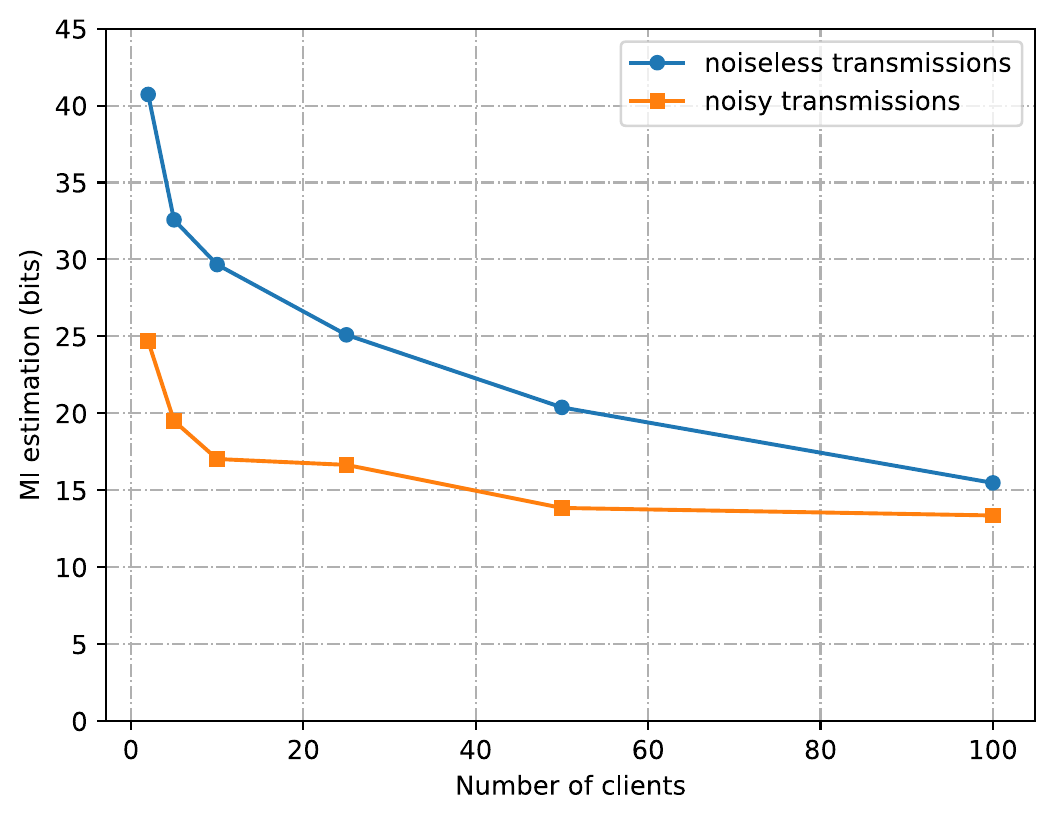}
    \caption{Impact of client number $N$ on privacy leakage, evaluated from training a CNN on the EMNIST dataset.}
    \label{fig:MI}
\end{figure}

\subsubsection{MI Estimation}
We use Mutual Information Neural Estimator (MINE) \cite{BelMohBar18mutual} to estimate the MI between each client's local gradient $\nabla \widetilde{f}_{n}(\boldsymbol{w}_t)$ and the aggregated global gradient $\boldsymbol{g}_t$. We follow similar procedures in \cite{elkordy22privacyFL} to obtain samples in the context of over-the-air computations. Specifically, we employ a fully connected neural network with two hidden layers, each containing 256 neurons, and using a learning rate of $\eta = 10^{-4}$. We perform 1000 global iterations to train the MINE network. For the noisy setting, we first sample $M$ sets of noise parameters $\{(\{c^{(m)}_{n,t}\}^{N}_{n=1};\boldsymbol{\xi}^{(m)}_t)\}^{M}_{m=1}$, and then loop these noise samples to calculate the aggregated model update for each gradient sample. This approach yields $M$ mutual information estimations for a single communication round, and we calculate the average of these estimates as the result. We evaluate the MI layer by layer and report the sum of the mutual information across all network layers as the final result.

%%------------------------------------------------%%
\subsection{Performance Evaluation}

\begin{figure}[t!]
    \centering    
	\begin{subfigure}{0.85\linewidth}
		\includegraphics[width=\linewidth]{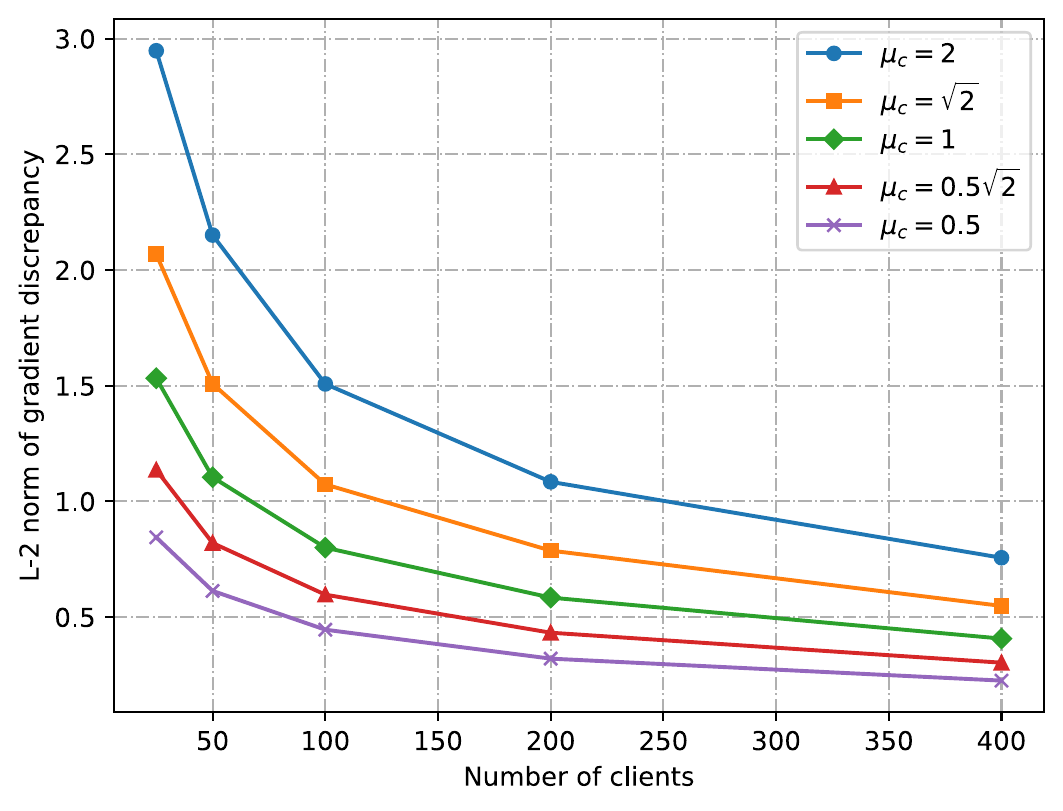}
		\caption{Rayleigh fading channel}	
        \label{fig:rayleigh_niid}
	\end{subfigure}

	\begin{subfigure}{0.85\linewidth}
		\includegraphics[width=\linewidth]{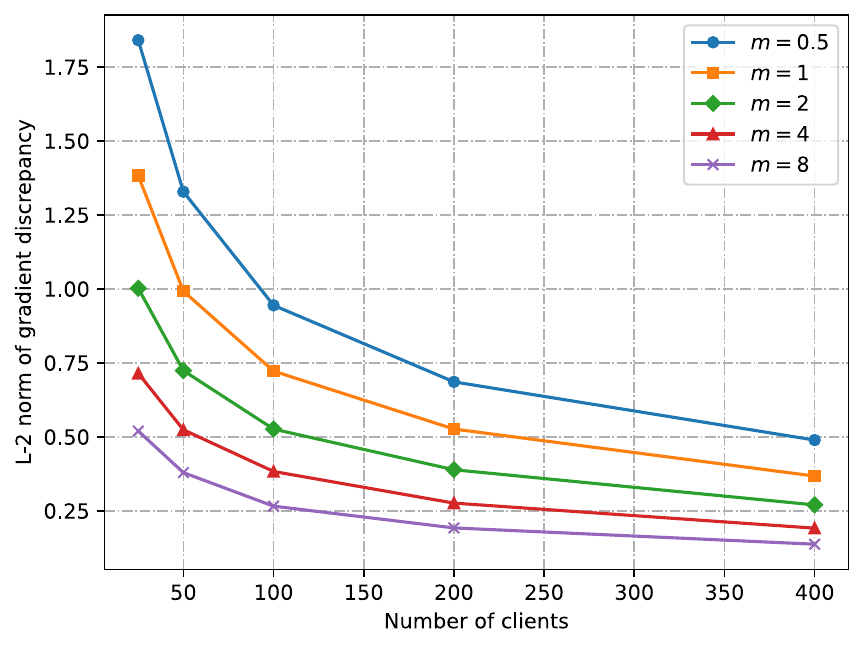}
		\caption{Nakagami fading channel}
        \label{fig:nakagami_niid}
	\end{subfigure}

    \caption{Visualizing the channel hardening effect, exemplified by training a CNN on the EMNIST dataset.}
    \label{fig:avg-eff}
\end{figure}

Fig.~\ref{fig:MI} draws the MI estimation as a function of the client number under noisy and noiseless transmission scenarios. Our experiments show that the estimated MI does not change significantly across communication rounds after the first few iterations. Hence, we average the estimates across training rounds when reporting the results. From this figure, we can see that as the number of clients increases, the privacy leakage of each participant decreases, which is consistent with our analysis in Theorem~\ref{theo:MI}. As such, it validates the benefits of system scaling up in enhancing privacy protection. Notably, the noisy setting results in a smaller MI for any given number of clients compared to the noiseless transmission scenario. This observation indicates that channel noise is an inherent encoding mechanism, offering natural privacy protection for over-the-air computing. The main reason can be attributed to the fact that the superposition property of multiple access channels effectively obscures individual client information within the globally aggregated gradient, whereas channel noise further perturbs the result. Moreover, as the number of clients increases, the challenge of distinguishing an individual local gradient from the aggregated gradient grows significantly, rendering it akin to finding a needle in a haystack.

In Fig.~\ref{fig:avg-eff}, we verify the channel hardening effect observed in Lemma~\ref{lma:Hoeffding_AggGrdt}. We use the actual channel coefficients $\{c_{n,t}\}^{n=N}_{n=1}$ and their mean $\mu_c$ to calculate a pair of aggregated gradients $\{\frac{1}{N}\sum_{n=1}^{N}c_{n,t}\nabla \widetilde{f}_{n}(\boldsymbol{w}_t);\frac{1}{N}\sum_{n=1}^{N}\mu_c\nabla \widetilde{f}_{n}(\boldsymbol{w}_t)\}$ at each communication round $t$. 
We then compute the L-2 norm of the discrepancy of the aggregated gradients pair and report the average across 200 global communication rounds. Note that we use the actual aggregated gradients $\frac{1}{N}\sum_{n=1}^{N}c_{n,t}\nabla \widetilde{f}_{n}(\boldsymbol{w}_t)$ to update the global model. We assess the impact of the number of clients under the Rayleigh and Nakagami fading channel and summarize the results in Fig.~\ref{fig:rayleigh_niid} and Fig.~\ref{fig:nakagami_niid}, respectively. The figures reveal that an increase in the number of participating clients leads to a decrease in the gradient discrepancy. This confirms that scaling up the system can mitigate the fluctuations in small-scale fading, driving the noisy global gradient close to the unperturbed version, effectively mitigating its impact, and enhancing the system's robustness.

\begin{figure}[t!]  
    \centering 
    \includegraphics[width=0.85\linewidth]
    % {Figures/batch_size_50_&_num_users_noise_niid_seed25_acc.svg}	
    {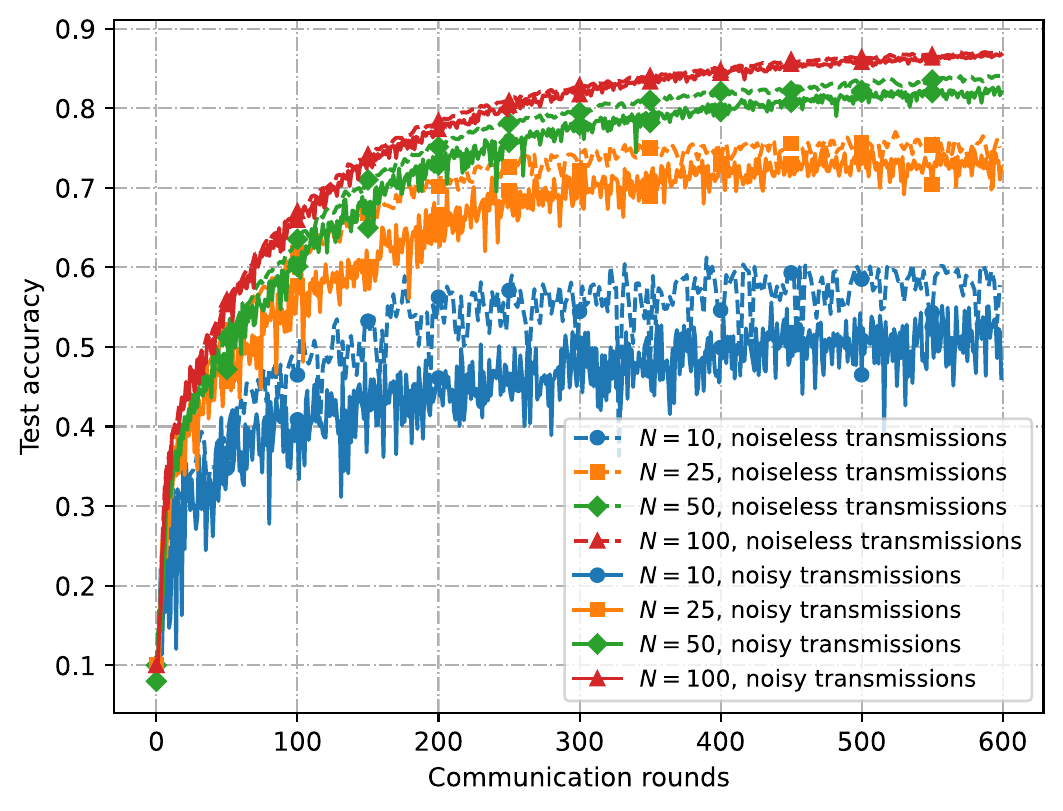}
    \caption{Impact of the number of clients $N$ on test accuracy, exemplified by training ResNet-18 on the CIFAR-10 dataset.}
    \label{fig:rate_n}
\end{figure}

In Fig.~\ref{fig:rate_n}, we investigate the test accuracy under a varying number of clients. The result shows that the test accuracy consistently improves with an increasing number of clients, demonstrating a positive influence from the enlarged scale of the system. Another noteworthy observation is that the convergence curve of over-the-air FL becomes smoother and approaches that under noiseless transmission as the number of clients increases, whereas it exhibits more fluctuations with fewer clients. 
These findings highlight the dual benefits of increasing the number of participating clients: It enhances the training performance by providing richer training datasets and reducing the gradient estimation and communication noise in model updates, thus leading to a more stable and swift convergence process.
As such, although blind transmission is not the optimal strategy from the power control aspect, it has a performance approaching the optimal in large-scale implementations of OTA FL.

We further explore the interplay between data heterogeneity and client number on the test accuracy in Fig.~\ref{fig:acc_hete}. In this figure, the Dirichlet distribution coefficient, denoted by $Dir$, is used to control the non-i.i.d. level of the data, where a lower $Dir$ value corresponds to a higher degree of heterogeneity. 
From Fig.~\ref{fig:acc_hete}, we can see that an increase in data heterogeneity impedes the training process, whereby it not only slows down the convergence rate but also inflicts additional fluctuations. However, with more clients participating in the system, the impact of data heterogeneity is reduced, and the model training experiences a more stable convergence performance. This reflects that scaling up the system can mitigate fluctuations caused by data heterogeneity, thereby improving the robustness of the training process.

% We also investigate the performance under different degrees of data heterogeneity. 
% The Dirichlet distribution coefficient, denoted as $Dir$, is used to control the non-i.i.d. nature of the data, where a lower $Dir$ value corresponds to a higher degree of heterogeneity.
% As illustrated in Fig.~\ref{fig:acc_hete}, the training performance slows down as the degree of data heterogeneity increases (i.e., as the value of $Dir$ decreases). 
% However, with more participants in the system, the convergence is less impacted by data heterogeneity and presents more stable performance.
% This reflects that scaling up the system can mitigate fluctuations caused by data heterogeneity, thereby improving the robustness of the training process.

\begin{figure}[t!]  
    \centering 
    \includegraphics[width=0.85\linewidth]{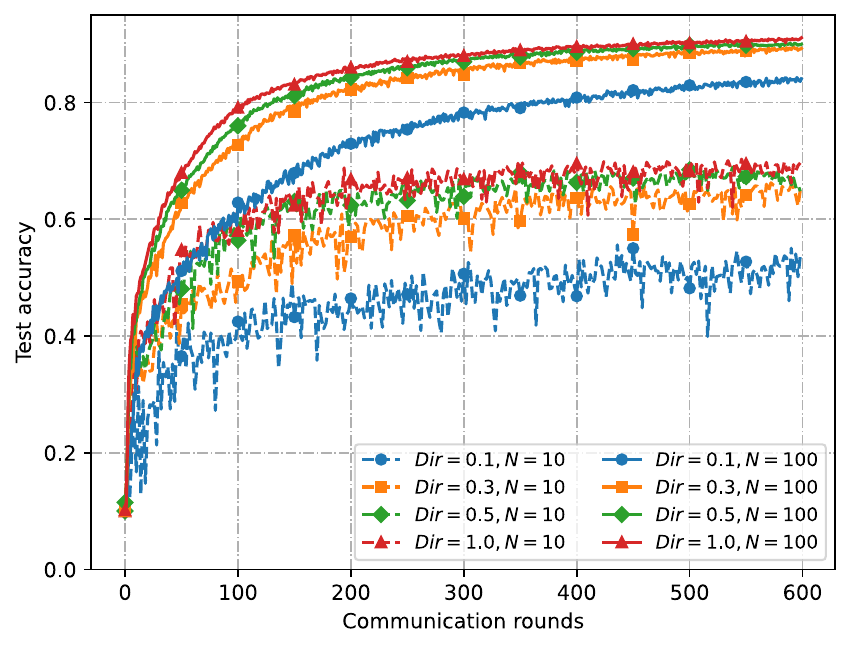}
    \caption{Impact of the number of clients $N$ on test accuracy under different levels of data heterogeneity $Dir$, exemplified by training ResNet-18 on the CIFAR-10 dataset.}
    \label{fig:acc_hete}
\end{figure}

\begin{figure}[t!]
    \centering 
    \includegraphics[width=0.85\linewidth]{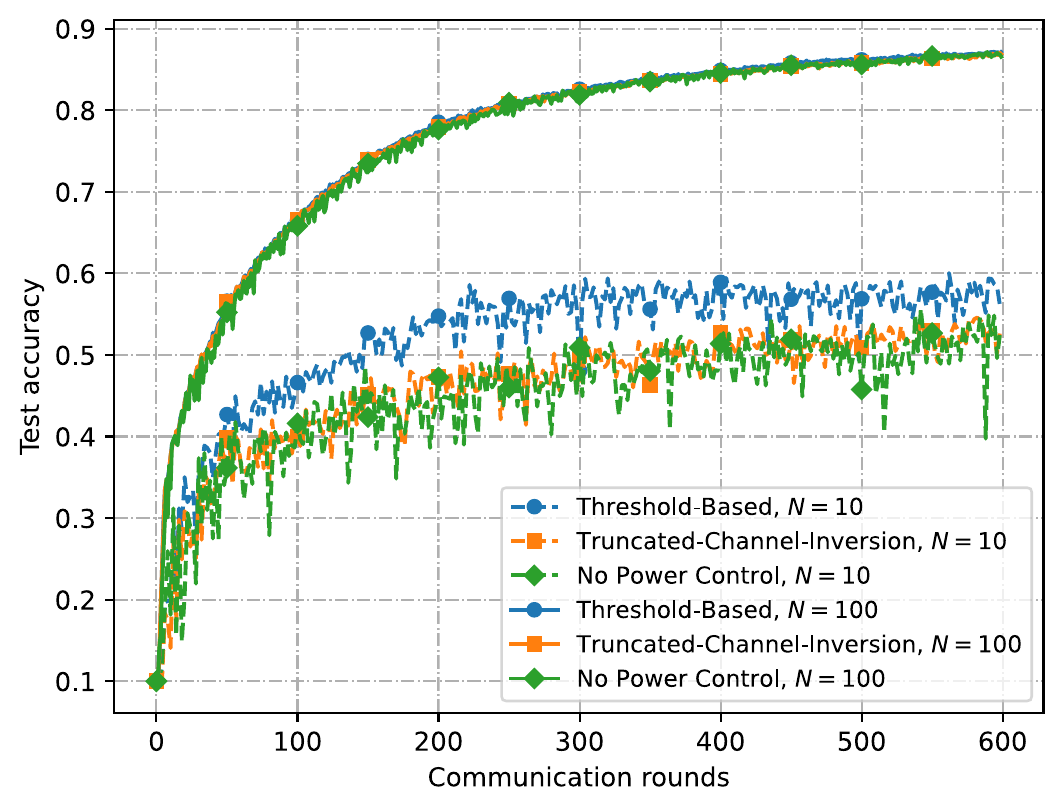}
    \caption{Impact of power control on training efficiency, exemplified by training ResNet-18 on the CIFAR-10 dataset.}
    \label{fig:rate_power}
\end{figure}

We examine the effectiveness of power control in Fig.~\ref{fig:rate_power} by comparing the test accuracy with and without power control. In this experiment, we adopt the threshold-based optimal power control \cite{cao20optimized} and the truncated channel inversion power control for comparison.
To mitigate aggregation information loss incurred by truncation operation, the cutoff threshold $c_{th}$ for channel inversion is set to satisfy $\mathbb{P}(c_{n,t}>c_{th})=0.99$. 
We assume perfect CSI and uniform instantaneous power constraints across all clients.
The figure shows that for a small number of clients, i.e., $N=10$, the threshold-based optimal power control exhibits superior performance, and both power control methods reduce fluctuations in the convergence curve.
However, when a large number of clients are present in the system, the convergence curve becomes smooth due to the channel hardening effect, and the performance gain from power control becomes almost negligible. This validates our observation in Section~III-D that instant CSI and the subsequent power control are not critical for over-the-air federated learning, particularly in large-scale systems.

\begin{figure}[t!]
    \centering    
	\begin{subfigure}{0.85\linewidth}
		\includegraphics[width=\linewidth]
        % {Figures/noisy_label_num_users_100_&_attack_portion_noise_niid_acc.svg}
        {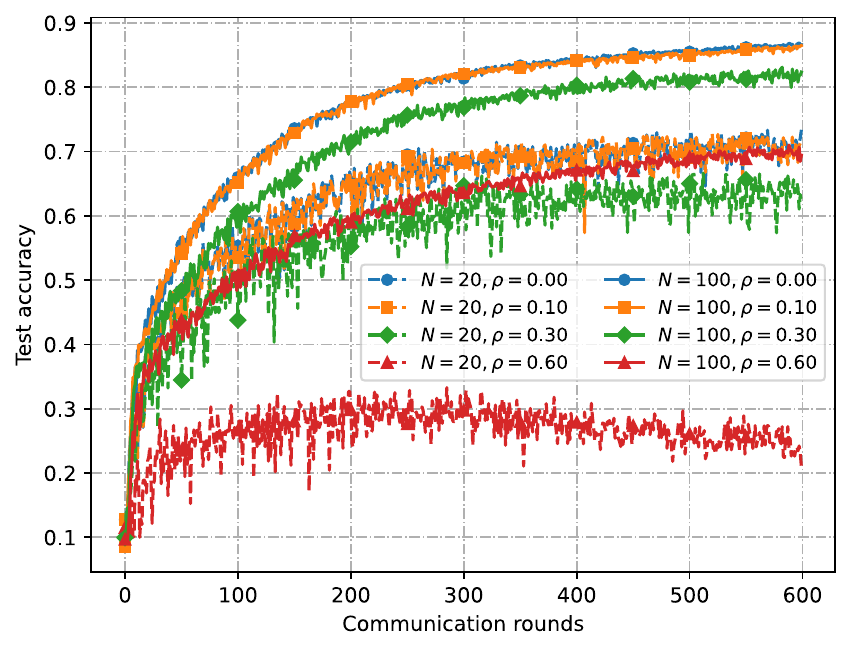}        
		\caption{noisy label attacks}	
        \label{fig:acc_att_noisy}
	\end{subfigure}
	\begin{subfigure}{0.85\linewidth}
		\includegraphics[width=\linewidth]
        % {Figures/class_flip_num_users_100_&_attack_portion_noise_niid_acc.svg}
        {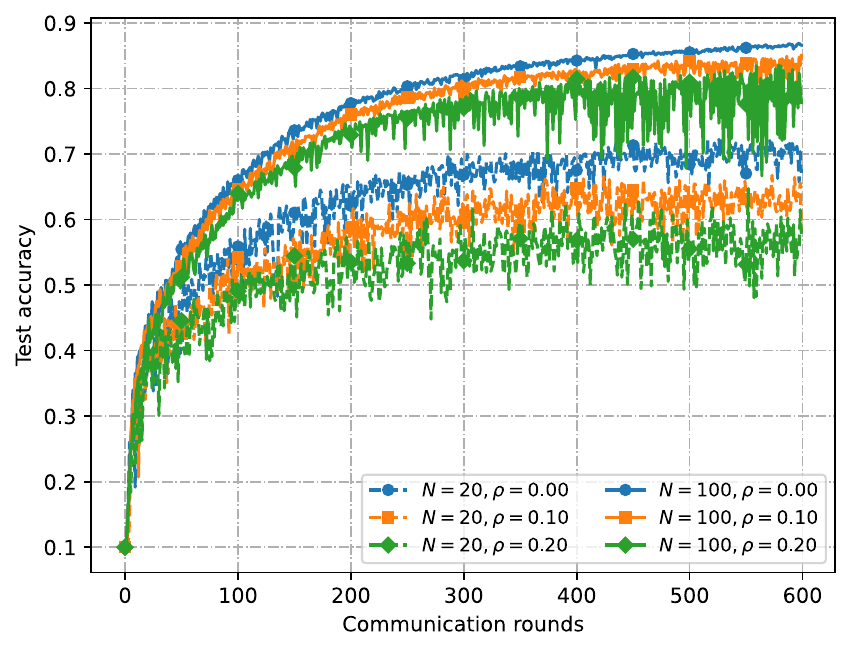}
		\caption{class flip attacks}
        \label{fig:acc_att_flip}
	\end{subfigure}
    \caption{Performance in the presence of malicious attacks, exemplified by training ResNet-18 on the CIFAR-10 dataset with $N=20$ and $N=100$.}
    \label{fig:rate_att}
\end{figure}

\subsection{Side Benefits of Scaling Up}

In this part, we examine the robustness of large-scale systems against malicious attacks.
Specifically, we consider two typical label attack patterns, i.e., the noisy label attack \cite{xu22fedcorr} and class flip attack \cite{yin18byzantine}.
We simulate the noisy label attacks by varying label noise levels at the malicious clients according to a uniform distribution. 
We alter the training labels at the malicious clients for the class flip attack by flipping each training label $i$ to $9-i$, specifically for the CIFAR-10 dataset.
The proportion of malicious clients is denoted by $\rho$, with a higher value of $\rho$ indicating a more severe overall attack on the system. 
Fig.~\ref{fig:acc_att_noisy} and Fig.~\ref{fig:acc_att_flip} display the test accuracy under these two attack types 
under different client numbers. 
% with $N=20$ and $N=100$ clients. 
The results indicate that while system performance degrades as attack intensity increases, the degradation is significantly less severe for the large-scale (i.e., $N=100$) system compared to the small-scale (i.e., $N=20$) system. 
When $N=100$, the convergence performance of the system remains stable even under relatively strong attack conditions (i.e., $\rho=0.6$). 
This highlights the benefit of scaling up the system, which enhances its resilience against malicious attacks.

\begin{figure}[t!]  
    \centering 
    \includegraphics[width=0.85\linewidth]{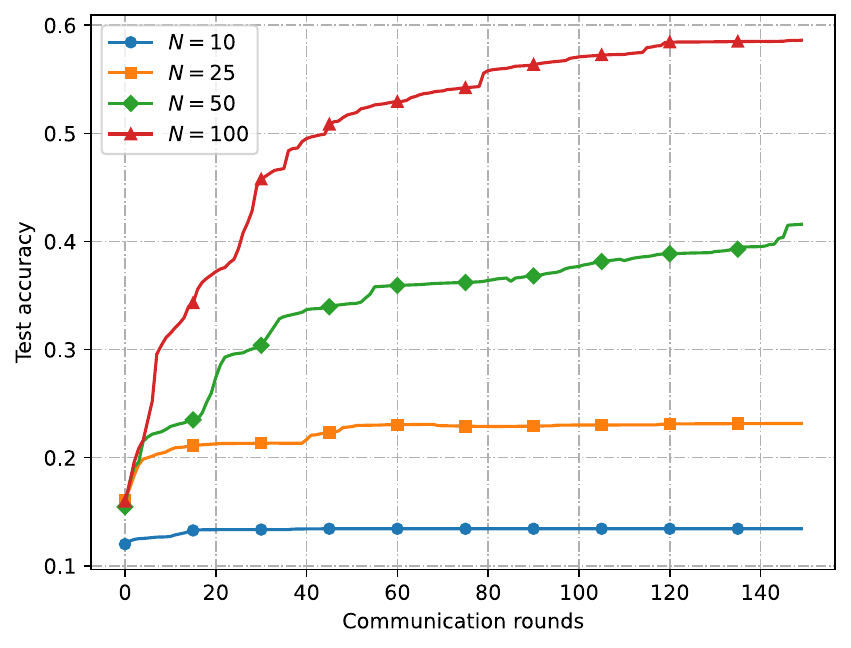}
    \caption{Impact of the number of clients $N$ on test accuracy, exemplified by training a logistic regression model on the EMNIST dataset.}
    \label{fig:acc_LBFGS}
\end{figure}

Finally, we focus on the learning methods employed at the edge server. In contrast to first-order methods that rely solely on gradient updates, second-order methods, such as the Newton method, achieve faster convergence by leveraging the Hessian matrix to refine model updates. However, higher-order methods are more vulnerable to parameter fluctuations and noise. We employ the Limited-memory Broyden Fletcher Goldfarb Shanno (L-BFGS) \cite{bonnans06numerical} algorithm, a representative quasi-Newton method, to update the global approximated inverse Hessian matrix at the edge server.
The distortion of the aggregated gradient compromises the accuracy of the second-order information derived from it, further exacerbating the challenge. To evaluate this, we train a logistic regression model for multi-class classification tasks on the EMNIST dataset and assess its performance across varying numbers of clients, as shown in Fig.~\ref{fig:acc_LBFGS}. The results demonstrate that increasing the number of clients significantly improves training performance, enabling the practical integration of second-order methods into over-the-air model aggregation. This improvement is attributed to system scaling, which effectively reduces the impact of communication noise on model updates, leading to a more stable convergence process.

%%%%%%%%%%%%%%%%%%%%%%%%%%%%%%%%%%%%%%%%%%%%%%%%%%%%
\section{Conclusion}
In this work, we carried out a theoretical study of the analog over-the-air model training in FL systems. We focus mainly on the system scale-up and evaluate its impact on privacy protection and training efficiency. Specifically, we have established an analytical expression for assessing privacy leakage in terms of mutual information during each round of global communication. We have also developed a Bernstein-like inequality that unveils a channel-hardening effect in analog over-the-air model aggregation. Subsequently, we have derived the convergence rate for both strongly convex and non-convex smooth loss functions in the setting of federated learning over the air. Our analysis reveals three benefits conferred by scaling up the system, namely, with an increasing number of clients: ($i$) the privacy leakage is substantially reduced since every client's gradient is obscured in the globally aggregated model, ($ii$) the channel hardening effect becomes more evident, which eliminates the impairments of small-scale fading, and ($iii$), the convergence rate can be accelerated as the thermal noise, gradient estimation noise, and the impact of data heterogeneity are further decreased. We have validated these theoretical findings by various experiments. Additionally, we have investigated the necessity of instant channel estimation and the subsequent power control in this setting. Our theoretical analysis and experimental findings confirm that instant power control is not essential for large-scale, over-the-air federated model training. Furthermore, the experimental results demonstrate that scaling up the system enhances resilience against malicious attacks and improves robustness in implementing second-order optimization methods.

While over-the-air FL systems enjoy substantial benefits from scaling up, they face the practical challenge of stringent synchronization requirements \cite{xiao24over}. 
To achieve time synchronization, existing studies have suggested techniques such as transmission timing control and broadcast clock sharing. 
The framework with a dual-layer synchronization interface proposed in \cite{pradhan2025experimental} has demonstrated successful synchronization in real-world experimental settings.
Ultimately, overcoming the synchronization challenge is crucial to ensure the reliable implementation of OTA-FL in practical wireless systems.

%%%%%%%%%%%%%%%%%%%%%%%%%%%%%%%%%%%%%%%%%%%%%%%%%%%%
\section{Appendix}
%%------------------------------------------------%%
\subsection{Proof of Theorem 1}
Without loss of generality, we derive the upper bound for the following term (as the change to the others only requires undergoing a permutation of client indices)
\begin{small}
\begin{align}
    I\!\left(\!\nabla \widetilde{f}_{N}(\boldsymbol{w}_t);\boldsymbol{g}_t \Big| \{ \boldsymbol{g}_{p} \}_{p\in[t-1]}\!\right).
\end{align}
\end{small}

Similar to \cite{elkordy22privacyFL}, we define $\nabla \bar{f_{n}}(\boldsymbol{w}_t)\in \mathbb{R}^{d^*_n}$, where $d^*_n\leq d$, which is a sub-vector with a rank-$d^*_n$ covariance matrix, induced from $\nabla \widetilde{f}_{n}(\boldsymbol{w}_t)$.
For ease of exposition, we denote by $\boldsymbol{\mu}_{n,t}$ and $\boldsymbol{K}_{n,t}$ the mean and covariance matrix of $\nabla \bar{f_{n}}(\boldsymbol{w}_t)$.
We also define $\boldsymbol{F}^{(t)}_{N}=\frac{1}{\sqrt{N}}\!\Big(\!\sum_{n=1}^{N}\!c_{n,t}\!\nabla \bar{f}_{n}(\boldsymbol{w}_t)\!+\!\bar{\boldsymbol{\xi}}_t \!\Big)$. 
To simplify the factors under consideration, we assume that $d^*_n = d^*$, $\forall n \in [N]$.
Then, we have
\begin{small}
\begin{align}\label{eq:MI_h}
    &I\!\left(\!\nabla \widetilde{f}_{N}(\boldsymbol{w}_t);\boldsymbol{g}_t \Big| \{ \boldsymbol{g}_{p} \}_{p\in[t-1]}\!\right) \notag \\
    &\overset{(a)}{=} I\!\left(\!\nabla \bar{f}_{N}(\boldsymbol{w}_t);\boldsymbol{F}^{(t)}_{N}\Big| \{ \boldsymbol{g}_{p} \}_{p\in[t-1]}\!\right) \notag \\
    % &=h\!\left(\!\nabla \bar{f}_{N}(\boldsymbol{w}_t)\Big| \{ \boldsymbol{g}_{p} \}_{p\in[t-1]}\!\right)
    % \!+\!h\left(\!\boldsymbol{F}^{(t)}_{N}\Big| \{ \boldsymbol{g}_{p} \}_{p\in[t-1]}\!\right) \notag \\
    %     &\qquad \qquad \qquad \qquad \qquad \qquad \quad -\!h\left(\nabla \bar{f}_{N}(\boldsymbol{w}_t),\boldsymbol{F}^{(t)}_{N}\Big| \{ \boldsymbol{g}_{p} \}_{p\in[t-1]}\right) \notag \\
    &=h\!\left(\!\nabla \bar{f}_{N}(\boldsymbol{w}_t)\Big| \{ \boldsymbol{g}_{p} \}_{p\in[t-1]}\!\right)
    \!+\!h\!\left(\!\boldsymbol{F}^{(t)}_{N}\Big| \{ \boldsymbol{g}_{p} \}_{p\in[t-1]}\!\right) \notag \\
        &\qquad \qquad \quad -\!h\!\left(\!\begin{bmatrix}\boldsymbol{I}_{d^*}&\boldsymbol{0}_{d^*}\\ \frac{c_{n,t}}{\sqrt{N}}\boldsymbol{I}_{d^*}&\frac{\sqrt{N\!-\!1}}{\sqrt{N}}\boldsymbol{I}_{d^*}\end{bmatrix}\!\!\begin{bmatrix}\nabla \bar{f}_{N}(\boldsymbol{w}_t)\\\boldsymbol{F}^{(t)}_{N-1}\end{bmatrix} \bigg| \{ \boldsymbol{g}_{p} \}_{p\in[t-1]}\!\right) \notag \\   
    &\overset{(b)}{=} h\!\left(\!\nabla \bar{f}_{N}(\boldsymbol{w}_t)\Big| \{ \boldsymbol{g}_{p} \}_{p\in[t-1]}\!\right)
    \!+\!h\!\left(\!\boldsymbol{F}^{(t)}_{N}\Big| \{ \boldsymbol{g}_{p} \}_{p\in[t-1]}\!\right) \notag \\
        &\quad -\!h\!\left(\!\nabla \bar{f}_{N}(\boldsymbol{w}_t)\Big| \{ \boldsymbol{g}_{p} \}_{p\in[t-1]}\!\right)
        \!-\!h\!\left(\!\boldsymbol{F}^{(t)}_{N-1}\Big| \{ \boldsymbol{g}_{p} \}_{p\in[t-1]}\!\right) \notag \\
        &\quad -\!\log{}{\!\Bigg|\det\!\begin{bmatrix}\boldsymbol{I}_{d^*}&\boldsymbol{0}_{d^*}\\ \frac{c_{n,t}}{\sqrt{N}}\boldsymbol{I}_{d^*}&\frac{\sqrt{N\!-\!1}}{\sqrt{N}}\boldsymbol{I}_{d^*}\end{bmatrix}\!\Bigg|} \notag \\
    &= h\!\left(\!\boldsymbol{F}^{(t)}_{N}\Big| \{ \boldsymbol{g}_{p} \}_{p\in[t-1]}\!\right) 
    \!-\!h\!\left(\!\boldsymbol{F}^{(t)}_{N-1}\Big| \{ \boldsymbol{g}_{p} \}_{p\in[t-1]}\!\right)\!+\!\frac{d^*}{2}\!\log{}{\!\Big(\frac{N}{N\!-\!1}\!\Big)}
\end{align}
\end{small}where $h(\cdot \vert \cdot)$ stands for the conditional entropy, $(a)$ holds because MI remains unchanged under multiplying with a constant, and $(b)$ follows from the property of the entropy of linear transformation of random vectors \cite{thomas06elements}. 

To further bound the first two terms in the last equality of \eqref{eq:MI_h}, we denote by $\boldsymbol{s}_{n,t}=c_{n,t}\nabla \bar{f}_{n} (\boldsymbol{w}_t)$.
Note that if $\boldsymbol{s}_{n,t}$ undergoes the following processing
% We preprocess $\boldsymbol{s}_{n,t}$ as follows
\begin{small}
\begin{align}
    \widehat{\boldsymbol{s}}_{n,t} = \boldsymbol{S}(\boldsymbol{s}_{n,t} - \mu_c \boldsymbol{\mu}_{n,t}) 
\end{align}
\end{small}where matrix $\boldsymbol{S}\in \mathbb{R}^{d^*\times d^*}$ is a diagonal matrix with each entry being i.i.d. drawn from the Bernoulli distribution $\mathcal{B}(0.5)$ with the support set $\{+1,-1\}$.
$\boldsymbol{S}$ is used for reducing the correlation among entries; the resultant vector $\widehat{\boldsymbol{s}}_{n,t}$ would have zero means, and its entries are mutually independent.
Therefore, its covariance matrix $\boldsymbol{\Sigma}_{n,t}$ can be represented as $\boldsymbol{\Sigma}_{n,t} = \mathrm{diag}(\sigma^2_{n,t,1}, \sigma^2_{n,t,2}, ..., \sigma^2_{n,t,d^*})$.
Then, we characterize the following entropy term for any $M \in \mathbb{N}$
\begin{small}
\begin{align}\label{eq:HM}
    &h \left( \boldsymbol{F}^{(t)}_{M}\Big| \{ \boldsymbol{g}_{p} \}_{p\in[t-1]} \right) =h \Bigg( \frac{1}{\sqrt{M}}\!\sum_{n=1}^{M}\!\boldsymbol{s}_{n,t} \!+\! \bar{\boldsymbol{\xi}_t} \Bigg| \{ \boldsymbol{g}_{p} \}_{p\in[t-1]} \Bigg) \notag \\
    &= h\!\Bigg(\!\frac{\boldsymbol{S}^{-1}}{\sqrt{M}}\!\Big( \sum_{n=1}^{M}  \widehat{\boldsymbol{s}}_{n,t}\!+\!\boldsymbol{S}\bar{\boldsymbol{\xi}_t}\!\Big)\Bigg| \{ \boldsymbol{g}_{p} \}_{p\in[t-1]}\!\Bigg) \notag \\
    &= \log{}{\left| \det \boldsymbol{S}^{-1}\right|} \!+\!\underbrace{h\!\Bigg(\!\frac{1}{\sqrt{M}}\Big(\!\sum_{n=1}^{M}\widehat{\boldsymbol{s}}_{n,t}\!+\!\boldsymbol{S}\bar{\boldsymbol{\xi}_t}\!\Big)\Bigg| \{ \boldsymbol{g}_{p} \}_{p\in[t-1]}\!\Bigg)}_{H_M}.
    % &= h\!\Bigg(\!\frac{\boldsymbol{S}^{-1}}{\sqrt{M}}\!\sum_{n=1}^{M}  \widehat{\boldsymbol{s}}_{n,t}\Bigg| \{ \boldsymbol{g}_{p} \}_{p\in[t-1]}\!\Bigg) \notag \\
    % &= \log{}{\left| \det \boldsymbol{S}^{-1}\right|} \!+\!\underbrace{h\!\Bigg(\!\frac{1}{\sqrt{M}}\sum_{n=1}^{M}\widehat{\boldsymbol{s}}_{n,t}\Bigg| \{ \boldsymbol{g}_{p} \}_{p\in[t-1]}\!\Bigg)}_{H_M}.
\end{align}
\end{small}

The upper bound of $H_M$ can be obtained by matching the first and second moments to a Gaussian-distributed vector, as follows: 
\begin{small}
\begin{align}\label{equ:HM_UppBnd}
    &H_M \leq \frac{1}{2} \log{}{ \left[ (2\pi e)^{d^*} \det \Big( \frac{1}{M}\!\sum_{n=1}^{M}\!\boldsymbol{\Sigma}_{n,t}\!+\!\frac{1}{M}\sigma^2_{z}\boldsymbol{I}_{d^*} \Big) \right]} \notag \\
    &\qquad  = \frac{1}{2}  \sum^{d^*}_{i=1}\log{}{ \left[2\pi e\Big( \frac{1}{M}\!\sum_{n=1}^{M}\!\sigma^2_{n,t,i} + \frac{\sigma^2_z}{M}\Big) \right]}.
    % &H_M \leq \frac{1}{2}\!\log{}{\!\left[\!(2\pi e)^{d^*}\!\det\!\Big(\!\frac{1}{M}\!\sum_{n=1}^{M}\!\boldsymbol{\Sigma}_{n,t}\!\Big)\!\right]} \notag \\
    % &\qquad  =\frac{1}{2}\!\sum^{d^*}_{i=1}\log{}{\!\left[2\pi e\Big(\!\frac{1}{M}\!\sum_{n=1}^{M}\!\sigma^2_{n,t,i}\!\Big)\!\right]}.
\end{align}
\end{small}And a lower bound of $H_M$ follows from leveraging the Berry-Esseen style bounds for the entropic central limit theorem \cite{bobkov14berry}: 
\begin{small}
\begin{align}\label{equ:HM_LowBnd}
    &H_M= \!\sum^{d^*}_{i=1} h \Bigg(  \frac{1}{ \sqrt{M}} \Big( \sum_{n=1}^{M}\!\widehat{\boldsymbol{s}}_{n,t,(i)}\!+\!\boldsymbol{S}\bar{\boldsymbol{\xi}}_{t,(i)} \Big)\Bigg| \{ \boldsymbol{g}_{p} \}_{p\in[t-1]} \Bigg) \notag \\   
    &\qquad \geq \frac{1}{2}\!\sum^{d^*}_{i=1}\log{}{ \left[2\pi e\Big( \frac{1}{M}\!\sum_{n=1}^{M}\!\sigma^2_{n,t,i}\!+\!\frac{\sigma^2_z}{M}\Big) \right]} - \frac{d^*C_{\bar{g}}}{M}
    % &H_M= \!\sum^{d^*}_{i=1}\!h\!\Bigg(\!\!\frac{1}{\!\sqrt{M}}\!\sum_{n=1}^{M}\!\widehat{\boldsymbol{s}}_{n,t}[i]\Bigg| \{ \boldsymbol{g}_{p} \}_{p\in[t-1]}\!\!\Bigg) \notag \\   
    % &\qquad \geq \frac{1}{2}\!\sum^{d^*}_{i=1}\log{}{\!\left[2\pi e\Big(\!\frac{1}{M}\!\sum_{n=1}^{M}\!\sigma^2_{n,t,i}\Big)\!\right]}\!-\!\frac{d^*C_{\bar{\boldsymbol{g}}}}{M}
\end{align}
\end{small}where $C_{\bar{\boldsymbol{g}}}$ denotes a specific constant associated with the finite fourth moment of $\widehat{\boldsymbol{s}}_{n,t,(i)}$.

Finally, by substituting \eqref{eq:HM} into \eqref{eq:MI_h}, we have
\begin{small}
\begin{align}\label{eq:MI_H}
    &I\!\left(\!\nabla \widetilde{f}_{N}(\boldsymbol{w}_t);\boldsymbol{g}_t\Big| \{ \boldsymbol{g}_{p} \}_{p\in[t-1]}\!\right)\!=\!H_N \!-\! H_{N\!-\!1} \!+\!\frac{d^*}{2}\! \log{}{\!\Big(\!\frac{N}{N\!-\!1}\!\Big)}.
\end{align}
\end{small}The proof is completed by substituting \eqref{equ:HM_UppBnd} (with $M=N$) and \eqref{equ:HM_LowBnd} (with $M=N-1$) into \eqref{eq:MI_H}.

\subsection{Proof of Lemma 1}
We define $\boldsymbol{z}_{n,t}=c_{n,t}\nabla \widetilde{f}_n ( \boldsymbol{w}_t )-\mu_c \nabla \widetilde{f}_n ( \boldsymbol{w}_t )$ for each $n \in [N]$.
Since $\{ c_{n,t} \}_{n=1}^N$ are i.i.d., it follows that $\mathbb{E}[\boldsymbol{z}_{n,t}] = \boldsymbol{0}$ and $\mathbb{E}[\|\boldsymbol{z}_{n,t,(i)} \|^2] \leq \mathbb{E}[\|\boldsymbol{z}_{n,t} \|^2] \leq \sigma^2_c G^2$, where $G = \max_{n \in [N]} (G_n)$.
Within this regime, we apply the extension of the vector Bernstein inequality \cite{kohler17sub-sampled} and obtain the following for $0 < \varepsilon < \sigma_c G$,
\begin{small}
\begin{align}
&\mathbb{P}\left( \bigg\Vert \frac{1}{N} \sum_{n=1}^N z_{n,t,(i)} \bigg\Vert \geq \varepsilon \right) \leq \exp\bigg(\!\! - N \frac{\varepsilon^2}{ 8 \sigma^2_c G^2 } \!+\! \frac{1}{4} \bigg).
\end{align}
\end{small}To handle the remaining regime, we apply Mcdiarmid’s inequality \cite{combes15extension}.
Specifically, define a function $F_{(i)}(\boldsymbol{c}_t)=\frac{1}{N} \sum^N_{n=1}c_{n,t}\nabla \widetilde{f}_{n,(i)} ( \boldsymbol{w}_t ), \forall i \in [d]$, with $\boldsymbol{c}_t=(c_{1,t}, ..., c_{N,t})$.
Given $\beta_\nu = \mathbb{P}( \vert c_{n,t} - \mu_c \vert > \nu )$, let $\mathcal{Y} \subset (\mathbb{R^+})^N$ be the subset where each entry of $\boldsymbol{c}_t \in \mathcal{Y}$ satisfies $|c_{n,t}-\mu_c| \leq \nu$.
Thus $F_{(i)}(\boldsymbol{c}_t)$ has a $\frac{2 \nu G}{N}$-bounded difference \cite{combes15extension} over $\mathcal{Y}$, and $\mathbb{P}(\boldsymbol{c}_t \notin \mathcal{Y}) = \beta_\nu^N$.
Therefore, for all $\varepsilon \geq 0$, we have
\begin{small}
\begin{align}
&\mathbb{P}\Big( \!F_i(\boldsymbol{c}_t) \!-\! \mathbb{E}\big[ F_i(\boldsymbol{c}_t)\big|\boldsymbol{c}_t \!\in\! \mathcal{Y} \big] \!\geq\! \varepsilon \!+\! 2\nu \beta_\nu^N G\Big) \!\leq\! \beta_\nu^N \!\!+\! \exp \!\Big(\!\! - \! \frac{ N \varepsilon^2 }{ 2 \nu^2 G^2 }\!\Big).
\end{align}
\end{small}

%%------------------------------------------------%%
\subsection{Proof of Theorem 2}
Similar to \cite{li19convergence}, we define a virtual sequence $\{\boldsymbol{v}^{(k)}_t\}$ to represent the averaged parameters at each iteration, where $\boldsymbol{v}^{(k)}_t=\frac{1}{N}\sum^N_{n=1}\boldsymbol{w}^{(k)}_{n,t}$.
For convenience, we denote by $\nabla \widetilde{f}_{n,t}^{(k)} = \frac{1}{B}\sum^{B}_{j=1}\nabla \ell(\boldsymbol{w}^{(k)}_{n,t};\theta^{(k)}_{n,t}(j))$ and $\nabla f_{n,t}^{(k)} = \frac{1}{M}\sum^{M}_{j=1}\nabla \ell(\boldsymbol{w}^{(k)}_{n,t};\theta^{(k)}_{n,t}(j))$.
Using the update rule, we have
\begin{small}
\begin{align}\label{eq:update}
    & \|\boldsymbol{v}^{(k+1)}_t \!\!-\! \boldsymbol{w}^* \|^2 = \bigg\|\boldsymbol{v}^{(k)}_t \!\!-\! \frac{\eta_l}{N}\!\!\sum^N_{n=1} \! \nabla \widetilde{f}_{n,t}^{(k)} \!-\! \boldsymbol{w}^* \bigg\|^2  \notag \\
    % &=\bigg\|\boldsymbol{v}^{(k)}_t \!\!-\! \frac{\eta_l}{N}\!\!\sum^N_{n=1} \! \nabla \widetilde{f}_{n,t}^{(k)} \!-\! \boldsymbol{w}^* \!-\! \frac{\eta_l}{N}\!\!\sum^N_{n=1} \! \nabla f_{n,t}^{(k)} \!+\! \frac{\eta_l}{N}\!\!\sum^N_{n=1} \! \nabla f_{n,t}^{(k)} \bigg\|^2 \notag \\
    &=\bigg\|\boldsymbol{v}^{(k)}_t \!\!-\! \frac{\eta_l}{N}\!\!\sum^N_{n=1} \! \nabla f_{n,t}^{(k)} \!-\! \boldsymbol{w}^* \bigg\|^2 \!\!+\! \eta_l^2 \bigg\|\frac{1}{N}\!\!\sum^N_{n=1} \! \nabla f_{n,t}^{(k)} \!\!-\! \frac{1}{N}\!\!\sum^N_{n=1} \! \nabla \widetilde{f}_{n,t}^{(k)} \bigg\|^2 \notag \\
    &\!+\! 2\eta_l \Big\langle \!\boldsymbol{v}^{(k)}_t \!\!-\! \frac{\eta_l}{N}\!\!\sum^N_{n=1} \! \nabla f_{n,t}^{(k)} \!-\! \boldsymbol{w}^* \!, \frac{1}{N}\!\!\sum^N_{n=1} \! \nabla f_{n,t}^{(k)} \!\!-\! \frac{1}{N}\!\!\sum^N_{n=1} \! \nabla \widetilde{f}_{n,t}^{(k)} \Big\rangle.
\end{align}
\end{small}Notice that the last term on the right hand side of \eqref{eq:update} has a zero mean. 
We follow the steps in \cite{li19convergence} and obtain the result of one step SGD, (if $\eta_l \leq \frac{1}{4L}$)
\begin{small}
\begin{align}\label{eq:one_step}
    &\mathbb{E}[\|\boldsymbol{v}^{(k+1)}_t \!\!-\! \boldsymbol{w}^* \|^2] 
    \leq \!(1 \!-\! \lambda \eta_l)\mathbb{E}[\|\boldsymbol{v}^{(k)}_t \!\!-\! \boldsymbol{w}^* \|^2] \!+\! 6L \eta_l^2 \Gamma  \notag \\   
    &\!+\! \eta_l^2 \mathbb{E}\bigg[\!\Big\| \!\frac{1}{N}\!\!\sum^N_{n=1} \! (\nabla f_{n,t}^{(k)} \!\!-\!\! \nabla \widetilde{f}_{n,t}^{(k)}) \!\Big\|^2 \!\bigg] \!\!+\! 2\mathbb{E} \bigg[\!\frac{1}{N}\!\!\sum^N_{n=1} \! \|\boldsymbol{v}^{(k)}_t \!\!-\! \boldsymbol{w}^{(k)}_{n,t} \|^2 \!\bigg].
\end{align}
\end{small}Leveraging \cite{khaled20tighter}, we have
\begin{small}
\begin{align}\label{eq:bound_v}
    \mathbb{E} \bigg[\frac{1}{N}\!\!\sum^N_{n=1} \! \|\boldsymbol{v}^{(k)}_t \!\!-\! \boldsymbol{w}^{(k)}_{n,t} \|^2 \bigg] \leq (E-1) \eta_l^2 \frac{\sum^N_{n=1} \sigma^2_{s,n}}{N B}.
\end{align}
\end{small}Hence, we can further bound the right-hand side of \eqref{eq:one_step} under Assumption \ref{assm:SGD sampling noise} as follows:
\begin{small}
\begin{align}\label{eq:one_step_f}
    &\mathbb{E}[\|\boldsymbol{v}^{(k+1)}_t \!\!-\! \boldsymbol{w}^* \|^2] 
    \leq (1 \!-\! \lambda \eta_l)\mathbb{E}[\|\boldsymbol{v}^{(k)}_t \!\!-\! \boldsymbol{w}^* \|^2] \!+\! 6L \eta_l^2 \Gamma  \notag \\   
    &\qquad \qquad \qquad \qquad \quad + \eta_l^2 \Big( \frac{1}{N} + 2(E\!-\!1) \Big) \frac{\sum^N_{n=1} \sigma^2_{s,n}}{N B}.
\end{align}
\end{small}We then apply \eqref{eq:one_step_f} recursively and arrive at the following:
\begin{small}
\begin{align}
    &\mathbb{E}[\|\boldsymbol{v}^{(E)}_t \!\!-\! \boldsymbol{w}^* \|^2] 
    \leq (1 \!-\! \lambda \eta_l)^E \|\boldsymbol{v}^{(0)}_t \!\!-\! \boldsymbol{w}^* \|^2  \notag \\  
    &\quad +\frac{1 \!-\! (1 \!-\! \lambda \eta_l)^E}{\lambda} \eta_l \bigg[6L \Gamma \!+\! \Big( \frac{1}{N} \!+\! 2(E\!-\!1) \Big) \frac{\sum^N_{n=1} \sigma^2_{s,n}}{N B} \bigg].
\end{align}
\end{small}Since $\eta_t=\frac{1}{\mu_c}\eta_l$, the global update yields:
\begin{small}
\begin{align}\label{eq:Cnvx_w}
    &\mathbb{E}[\|\boldsymbol{w}_{t+1} \!-\! \boldsymbol{w}^* \|^2] = \mathbb{E}\bigg[\!\Big\|\boldsymbol{w}_t \!-\! \frac{\eta_t}{N}\!\Big(\!\sum^N_{n=1}\!\mu_c \frac{\boldsymbol{w}_t\!-\!\boldsymbol{w}^{(E)}_{n,t}}{\eta_l} \!+\! \boldsymbol{\xi}_t \!\Big)\!-\!\boldsymbol{w}^* \Big\|^2 \bigg]  \notag \\ 
    &=\mathbb{E}\bigg[\!\Big\|\frac{1}{N}\!\!\sum^N_{n=1}\boldsymbol{w}^{(E)}_{n,t} \!-\!\boldsymbol{w}^* \!-\! \frac{\eta_t}{N} \boldsymbol{\xi}_t \Big\|^2 \bigg] = \mathbb{E}[\!\|\boldsymbol{v}^{(E)}_t \!-\!\boldsymbol{w}^*\|^2 ] \!+\! \eta_t^2 \frac{d\sigma^2_z}{N^2} \notag \\ 
    &\leq (1 \!-\! \lambda \eta_l)^E\mathbb{E}[\|\boldsymbol{w}_t \!-\! \boldsymbol{w}^* \|^2] + \eta_l^2 \frac{d\sigma^2_z}{\mu_c^2 N^2} \notag \\ 
    &\quad + \frac{1 \!-\! (1 \!-\! \lambda \eta_l)^E}{\lambda} \eta_l \bigg[6L \Gamma \!+\! \Big( \frac{1}{N} \!+\! 2(E\!-\!1) \Big) \frac{\sum^N_{n=1} \sigma^2_{s,n}}{N B} \bigg].
\end{align}
\end{small}We complete the proof by simple algebra and removing the high-order terms as they become infinitesmall as $t$ goes large.

%%------------------------------------------------%%
\subsection{Proof of Corollary 1}
We rewrite \eqref{eq:Cnvx_w} as follows by invoking Bernoulli's inequality:
% Applying the Bernoulli's inequality, we can rewrite \eqref{eq:Cnvx_w} as follows
\begin{small}
\begin{align}
    &\mathbb{E}[\|\boldsymbol{w}_{t+1} \!-\! \boldsymbol{w}^* \|^2] 
    \leq (1 \!-\! \lambda \eta_l) \mathbb{E}[\|\boldsymbol{w}_t \!-\! \boldsymbol{w}^* \|^2]   \notag \\
    &\quad + \eta_l^2 \underbrace{\left[ \frac{d\sigma^2_z}{\mu_c^2 N^2} \!+\! E \bigg[6L \Gamma \!+\! \Big( \frac{1}{N} \!+\! 2(E\!-\!1) \Big) \frac{\sum^N_{n=1} \sigma^2_{s,n}}{N B} \bigg] \right]}_{Q_1}.
\end{align}
\end{small}The proof proceeds by induction.
Firstly, the definition of $\kappa$ ensures that it holds for the initial case $t=0$.
Then, assuming the bound holds for some $t$ and applying the diminishing learning rate $\eta_t = \frac{\eta_{0}}{1+t}$, where $\eta_{0} \leq \frac{1}{4 \mu_c L}$, it follows that
\begin{small}
\begin{align}
    &\mathbb{E}[\|\boldsymbol{w}_{t+1} \!-\! \boldsymbol{w}^* \|^2] 
    \leq (1 \!-\! \frac{\lambda \mu_c \eta_0}{1+t})\frac{\kappa}{1+t} + \frac{\mu_c^2 \eta_0^2}{(1+t)^2} Q_1  \notag \\
    &\qquad \quad =\frac{t}{(1+t)^2} \kappa + \bigg[ \frac{\mu_c^2 \eta_0^2 Q_1}{(1+t)^2} - \frac{\lambda \mu_c \eta_0 \!-\! 1}{(1+t)^2} \kappa \bigg]\leq \frac{\kappa}{2+t}.
\end{align}
\end{small}Thus, the bound holds for $t+1$, completing the proof.

%%------------------------------------------------%%
\subsection{Proof of Theorem 3}
For ease of exposition, let us define $\widetilde{\boldsymbol{g}}_t=\frac{1}{\mu_c E}\boldsymbol{g}_t$.
Then, using the smoothness of $f(\boldsymbol{w})$, we have
\begin{small}
\begin{align}\label{eq:L-smoothness}
    &\mathbb{E}[f\!(\!\boldsymbol{w}_{t+1}\!)]\!-\!\mathbb{E}[f\!(\!\boldsymbol{w}_t\!)] \!\leq\!\mathbb{E}\langle \nabla\! f(\boldsymbol{w}_t), \!\boldsymbol{w}_{t+1}\!-\!\!\boldsymbol{w}_t \rangle 
    \!+\!\frac{L}{2}\mathbb{E}[\|\boldsymbol{w}_{t+1}\!-\!\!\boldsymbol{w}_t\|^2] \notag \\    
    &=\!- \frac{\eta_t E\mu_c}{2} \mathbb{E}[\| \nabla \!f(\boldsymbol{w}_t) \|^2]
    \!-\! \frac{\eta_t E\mu_c(1\!-\!\eta_t E \mu_c L)}{2}\mathbb{E}[\|\widetilde{\boldsymbol{g}}_t\|^2] \notag \\
    &\qquad \qquad \qquad \qquad \qquad \qquad +\! \frac{\eta_t E\mu_c}{2}\mathbb{E}[\| \nabla \!f(\boldsymbol{w}_t)\!-\!\widetilde{\boldsymbol{g}}_t \|^2]. 
\end{align}    
\end{small}

By noticing that $\mathbb{E}[\boldsymbol{\xi}_t]=\boldsymbol{0}$ and applying Assumption \ref{assm:gradient bound}, we can bound $\mathbb{E}[\|\widetilde{\boldsymbol{g}}_t\|^2]$ as follows:
\begin{small}
\begin{align}
    &\mathbb{E}[\|\widetilde{\boldsymbol{g}}_t\|^2]
    =\mathbb{E}\!\left[\!\bigg\|\!\frac{1}{E}\!\Big(\!
    \frac{1}{N}\!\!\sum_{n=1}^{N}\!\!\sum^{E-1}_{k=0}\!\!\nabla \widetilde{f}_{n,t}^{(k)}\!+\!\frac{\boldsymbol{\xi}_t}{\mu_c N}\!\Big)\!\bigg\|^2\!\right]\notag \\
    &\!=\!\mathbb{E}\!\!\left[\!\bigg\|\!\frac{1}{N}\!\!\sum_{n=1}^{N}\!\!\frac{1}{E}\!\!\sum^{E-1}_{k=0}\!\!\nabla \widetilde{f}_{n,t}^{(k)}\bigg\|^2\!\right] \!\!+\!\mathbb{E}\!\!\left[\!\bigg\|\!\frac{\boldsymbol{\xi}_t}{\mu_c EN}\bigg\|^2\!\right] 
    \!\!\leq \! \frac{1}{N}\!\!\sum^N_{n=1}\! G_n^2 \!+\! \frac{d\sigma_{z}^2}{\mu_c^2 E^2 N^2}.    
\end{align}
\end{small}Next, we expand $\mathbb{E}[\|\nabla f(\boldsymbol{w}_t)-\widetilde{\boldsymbol{g}}_t\|^2]$ as follows:
\begin{small}
\begin{align}
    &\mathbb{E}[\|\nabla f(\boldsymbol{w}_t)-\widetilde{\boldsymbol{g}}_t\|^2] \notag \\
    &=\mathbb{E}\!\left[\!\bigg\|\!\frac{1}{N}\!\!\sum_{n=1}^{N}\!\nabla f_{n,t}^{(0)} \!-\!\frac{1}{E}\!\Big(
    \frac{1}{N}\!\!\sum_{n=1}^{N}\!\!\sum^{E-1}_{k=0}\!\nabla \widetilde{f}_{n,t}^{(k)}\!+\!\frac{\boldsymbol{\xi}_t}{\mu_c N}\!\Big)\!\bigg\|^2\right] \notag \\
    &=\mathbb{E}\!\Bigg[\!\bigg\|\frac{1}{N}\!\!\sum_{n=1}^{N}\underbrace{\!\Big(\!\frac{1}{E}\!\sum^{E-1}_{k=0}\!\nabla \widetilde{f}_{n,t}^{(k)} \!-\! \nabla f_{n,t}^{(0)}\Big)}_{\boldsymbol{e}_{n,t}}\bigg\|^2\Bigg]
    \!+\!\mathbb{E}\!\left[\!\bigg\|\!\frac{\boldsymbol{\xi}_t}{\mu_c EN}\bigg\|^2\right]      
\end{align}
\end{small}where $\boldsymbol{e}_{n,t}$ can be further decomposed into the following:
\begin{small}
\begin{align}
    &\boldsymbol{e}_{n,t}\!=\!\underbrace{\frac{1}{E}\!\!\sum^{E-1}_{k=0}\!\!\Big(\nabla \widetilde{f}_{n,t}^{(k)}\!-\!\nabla f_{n,t}^{(k)} \Big)}_{\hat{\boldsymbol{e}}_{n,t}}\!+\!\underbrace{\frac{1}{E}\!\!\sum^{E-1}_{k=0}\!\!\Big(\nabla f_{n,t}^{(k)}\!-\!\nabla f_{n,t}^{(0)}\Big)}_{\bar{\boldsymbol{e}}_{n,t}}.
\end{align}
\end{small}

Subsequently, we bound $\hat{\boldsymbol{e}}_{n,t}$ and $\bar{\boldsymbol{e}}_{n,t}$, respectively.
% Here $\hat{\boldsymbol{e}}_{n,t}$ is related to the accumulative SGD sampling noise.
According to Assumption~\ref{assm:SGD sampling noise}, we have $\mathbb{E}[\hat{\boldsymbol{e}}_{n,t}]=\boldsymbol{0}$, hence
\begin{small}
\begin{align}
    &\mathbb{E}\left[\bigg\|\frac{1}{N}\!\!\sum^N_{n=1}\!\hat{\boldsymbol{e}}_{n,t}\bigg\|^2\right]
    = \frac{1}{N^2}\!\!\sum^N_{n=1}\!\frac{1}{E^2}\!\!\sum^{E-1}_{k=0}\!\mathbb{E}\!\left[\!\Big\|\nabla \widetilde{f}_{n,t}^{(k)} \!-\!\nabla f_{n,t}^{(k)}\Big\|^2\right]  \notag \\
    &\qquad \qquad \qquad \qquad \leq \frac{\sum^{N}_{n=1}\!\sigma^2_{s,n}}{N^2\!B E}.
\end{align}
\end{small}On the other hand, using Assumption~\ref{assm:gradient bound} and smoothness of the loss function, we can bound $\bar{\boldsymbol{e}}_{n,t}$ via the following:
\begin{small}
\begin{align}
    &\mathbb{E}[\|\bar{\boldsymbol{e}}_{n,t}\|^2]
    \!\leq\!\frac{1}{E}\!\!\sum^{E-1}_{k=0}\!\mathbb{E}\!\left[\!\Big\|\!\nabla f_{n,t}^{(k)}\!-\!\nabla f_{n,t}^{(0)}\!\Big\|^2\!\right]
    \!\leq\! \frac{L^2}{E}\!\sum^{E-1}_{k=0}\!\mathbb{E}[\|\boldsymbol{w}^{(k)}_{n,t}\!-\!\boldsymbol{w}_t\|^2]\notag \\
    &= \frac{L^2}{E}\!\!\sum^{E-1}_{k=0}\!\mathbb{E}\!\Bigg[\bigg\|\eta_l\!\sum^{k-1}_{p=0}\!\nabla \widetilde{f}_{n,t}^{(p)}\bigg\|^2\Bigg]
    \!\leq\! \frac{L^2{\eta_l}^2 G_n^2(E\!-\!1)(2E\!-\!1)}{6}.
\end{align}  
\end{small}

To this end, we obtain
\begin{small}
\begin{align}\label{eq:bound_f-g}
    &\mathbb{E}[\|\nabla f(\boldsymbol{w}_t)-\widetilde{\boldsymbol{g}}_t\|^2]=
    \mathbb{E}\!\Bigg[\bigg\|\frac{1}{N}\!\sum_{n=1}^{N}\!\boldsymbol{e}_{n,t}\bigg\|^2\Bigg] \!+\!\mathbb{E}\!\left[\bigg\|\!\frac{\boldsymbol{\xi}_t}{\mu_c EN}\bigg\|^2\right] \notag \\
    &= \mathbb{E}\!\left[\bigg\|\frac{1}{N}\!\sum^N_{n=1}\!\hat{\boldsymbol{e}}_{n,t}\bigg\|^2\right] \!+\! \mathbb{E}\!\left[\bigg\|\frac{1}{N}\!\sum^N_{n=1}\!\bar{\boldsymbol{e}}_{n,t}\bigg\|^2\right] \!+\! \mathbb{E}\!\left[\bigg\|\!\frac{\boldsymbol{\xi}_t}{\mu_c EN}\bigg\|^2\right] \notag \\
    &\leq \frac{\sum^N_{n=1}\sigma^2_{s,n}}{N^2\!B E}\!+\!\frac{L^2{\eta_l}^2 (E\!-\!1)(2E\!-\!1)\!\sum^N_{n-1}\!G_n^2}{6N} \!+\! \frac{d\sigma^2_z}{\mu_c^2 E^2 N^2}.
\end{align}
\end{small} 

Since $\eta_t=\frac{1}{\mu_c}\eta_l=\frac{1}{\mu_c L}$, by simple algebra, we have
\begin{small}
\begin{align}\label{eq:non-convex}
    &\mathbb{E}[f(\boldsymbol{w}_{t+1})] \!-\! f(\boldsymbol{w}^*)\leq\mathbb{E}[f(\boldsymbol{w}_t)]\!-\!f(\boldsymbol{w}^*)\!-\!\frac{E}{2L}\mathbb{E}[\|\nabla f(\boldsymbol{w}_t)\|^2]\notag \\
    &\qquad \qquad +\!\frac{d\sigma_{z}^2}{2\mu_c^2 N^2 L}\!+\!\frac{\sum^N_{n=1}\sigma^2_{s,n}}{2N^2\!B\!L}\!+\!\frac{E(E\!-\!1)(2E\!+\!5)\sum^N_{n=1}\!G_n^2}{12NL}.
\end{align}
\end{small}By summing up \eqref{eq:non-convex} for $t = 0,1,\cdots, T\!-\!1$ and rearranging the above formula, we complete the proof.

%%------------------------------------------------%%
\subsection{Proof of Corollary 2}
Substituting $\|\widetilde{\boldsymbol{g}}_t\|^2 \leq 2\|\nabla f(\boldsymbol{w}_t) \!-\! \widetilde{\boldsymbol{g}}_t \|^2 \!+\! 2\|\nabla f(\boldsymbol{w}_t) \|^2$ into \eqref{eq:L-smoothness}, we have
\begin{small}
\begin{align}
    &\mathbb{E}[f(\!\boldsymbol{w}_{t+1}\!)]\!-\!\mathbb{E}[f(\!\boldsymbol{w}_t\!)] \!\leq\! \eta_t E\mu_c (\eta_t E \mu_c L -\frac{3}{2})\mathbb{E}[\| \nabla \!f(\!\boldsymbol{w}_t\!) \|^2] \notag \\
    &\qquad \qquad \qquad  + \eta_t E\mu_c (\eta_t E \mu_c L -\frac{1}{2}) \mathbb{E}[\| \nabla \!f(\!\boldsymbol{w}_t\!)\!-\!\!\widetilde{\boldsymbol{g}}_t \|^2]. 
\end{align}    
\end{small}Since $\eta_t = \frac{1}{\mu_c}\eta_l = \frac{\eta_0}{1+t}$ and $\eta_0 < \frac{1}{\mu_c EL}$, we obtain
\begin{small}
\begin{align}\label{eq:non-convex_decay}
    &\mathbb{E}[f(\!\boldsymbol{w}_{t+1}\!)]\!-\!\mathbb{E}[f(\!\boldsymbol{w}_t\!)] \leq -\frac{\eta_t E\mu_c}{2} \mathbb{E}[\| \nabla \!f(\!\boldsymbol{w}_t\!) \|^2]  \notag \\
    &\qquad \qquad \qquad \qquad \qquad \quad \!\!+ \eta_t^2 E ^2 \mu_c^2 L \mathbb{E}[\| \nabla \!f(\!\boldsymbol{w}_t\!)\!-\!\!\widetilde{\boldsymbol{g}}_t \|^2]. 
\end{align}    
\end{small}Adopting \eqref{eq:bound_f-g} and summing \eqref{eq:non-convex_decay} over $t = 0,1,\cdots,T\!-\!1$, we rearrange the terms to obtain the following result:
\begin{small}
\begin{align}
    &\sum^{T-1}_{t=0}\!\frac{\mathbb{E}[\| \nabla \!f(\!\boldsymbol{w}_t\!) \|^2]}{1 \!+\! t} \!\leq\! \frac{2\left(f(\boldsymbol{w}_0) \!-\! f(\boldsymbol{w}^*)\right)}{\eta_0 E \mu_c} \!+\! \bigg[\! \frac{\sum^N_{n=1}\sigma^2_{s,n}}{N^2 B E}\!+\! \frac{d \sigma^2_z}{\mu_c^2 E^2 N^2} \!\bigg]   \notag \\
    &\times \sum^{T \!-\!1}_{t=0} \!\frac{2 \eta_0 E \mu_c L}{(1 \!+\! t)^2} \!+\! \frac{\eta_0^3 \mu_c^3 L^3 E(E \!-\! 1)(2E \!-\! 1)\!\sum^N_{n=1}\!G_n^2}{3N} \!\sum^{T \!-\! 1}_{t=0} \!\frac{1}{(1 \!+\! t)^4}.
\end{align}    
\end{small}Invoking the bound $\sum^{T-1}_{t=0} \frac{1}{1+t} > \log{}{T}$, $\sum^{T-1}_{t=0} \frac{1}{(1+t)^2} < \frac{\pi^2}{6}$, and $\sum^{T-1}_{t=0} \frac{1}{(1+t)^4} < \frac{\pi^4}{90}$ gives the result.

%%%%%%%%%%%%%%%%%%%%%%%%%%%%%%%%%%%%%%%%%%%%%%%%%%%%
%\balance
\bibliographystyle{IEEEtran}
\bibliography{bib/StringDefinitions,bib/IEEEabrv,bib/A-OTA_scale_up}

\end{document}